\documentclass[10pt,twocolumn,letterpaper]{article}

 \usepackage{cvpr}              

\usepackage{booktabs}
\usepackage{amsfonts}
\usepackage{nicefrac}
\usepackage{microtype}
\usepackage{graphicx}
\usepackage{array}
\usepackage{fvextra}
\usepackage{amsmath}
\usepackage[export]{adjustbox}
\usepackage{caption}
\usepackage{float}
\usepackage{adjustbox}
\definecolor{cvprblue}{rgb}{0.21,0.49,0.74}
\usepackage[pagebackref,breaklinks,colorlinks,allcolors=cvprblue]{hyperref}

\def\confName{CVPR}
\def\confYear{2027}

\title{CheXTemporal: A Dataset for Temporally--Grounded Reasoning in Chest Radiography}

\author{
Eva Prakash\textsuperscript{1}\thanks{Corresponding author.}
\enspace
Yunhe Gao\textsuperscript{1}
\enspace
Chong Wang\textsuperscript{1}
\enspace
Neal Prakash\textsuperscript{2}
\enspace
Justin Xu\textsuperscript{3}\\
Arne Edward Michalson\textsuperscript{1}
\enspace
Seena Dehkharghani\textsuperscript{1}
\enspace
Eun Kyoung Hong\textsuperscript{1}
\enspace
Julie Bauml\textsuperscript{4}\\
Roger Boodoo\textsuperscript{5}
\enspace
Sophie Ostmeier\textsuperscript{6}
\enspace
Christian Bluethgen\textsuperscript{1}
\enspace
Jean-Benoit Delbrouck\textsuperscript{5}
\enspace
Curtis Langlotz\textsuperscript{1}\\
{\small
\textsuperscript{1}Stanford University
\quad
\textsuperscript{2}Columbia University
\quad
\textsuperscript{3}University of Oxford}\\
{\small
\textsuperscript{4}Vanderbilt University
\quad
\textsuperscript{5}HOPPR
\quad
\textsuperscript{6}ETH Zurich}\\
{\tt\small eprakash@stanford.edu}
}

\begin{document}
\maketitle
\begin{abstract}
Chest radiograph interpretation requires temporal reasoning over prior and current studies, yet many vision--language models are trained on static image--report pairs and lack explicit supervision for modeling longitudinal change. We introduce CheXTemporal, a dataset for temporally grounded reasoning in chest radiography consisting of paired prior--current chest X-rays (CXRs) with finding--level temporal and spatial annotations. The dataset includes a five--class progression taxonomy (new, worse, stable, improved, resolved), localized spatial supervision, explicit spatial--temporal alignment, and multi--source coverage. In addition to our gold--standard evaluation set, we construct a large--scale silver dataset with automatically derived temporal and anatomical supervision for training progression--aware representations. Using these resources, we evaluate state--of--the--art vision--language CXR models on zero--shot progression classification. We further introduce Prog-JEPA, a progression-aware joint-embedding predictive framework built on VL-JEPA and trained on CheXTemporal silver annotations, and investigate alternative formulations of temporal supervision. Prog-JEPA achieves stronger progression classification performance than a supervised baseline and pretrained vision--language models on external benchmarks, while improving progression-specific spatial grounding under controlled text prompts.
\end{abstract}

\section{Introduction}

\begin{figure*}[t]
    \centering
    \includegraphics[width=0.8\textwidth]{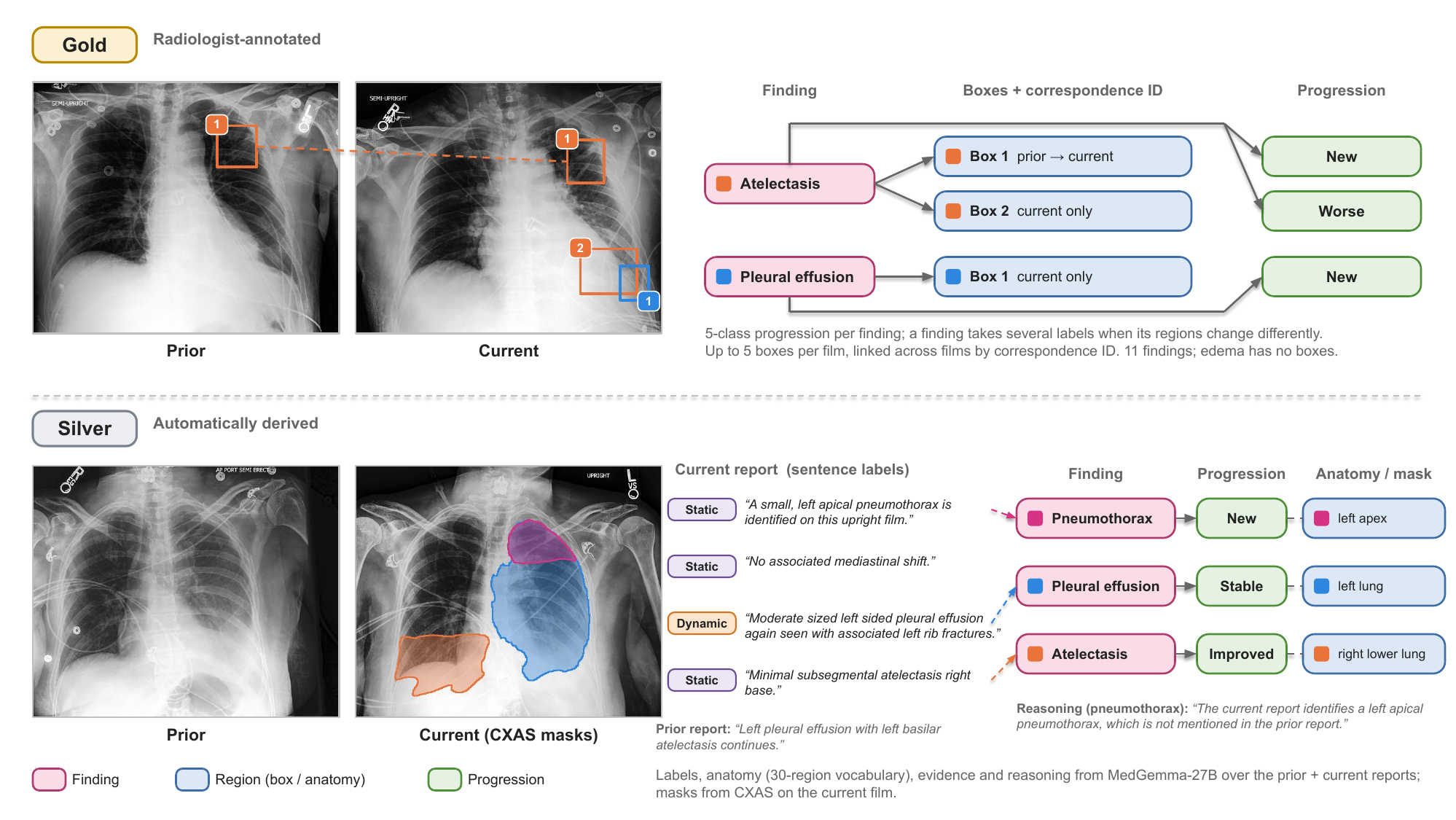}
    \vspace{-10pt}
    \caption{Overview of CheXTemporal gold and silver annotations. The gold set contains radiologist-assessed progression labels, pathology bounding boxes on paired prior and current CXRs, and cross-time region correspondence with region-specific progression labels. A finding may worsen in one region while newly appearing in another. The silver set provides large-scale report-derived progression labels, supporting evidence sentences, anatomical locations, and spatial masks.}
    \label{fig:chextemporal_annotations}
\end{figure*}

Longitudinal comparison is central to chest radiography. Radiologists routinely examine current and prior chest X-rays (CXRs) to determine whether abnormalities have newly appeared, worsened, remained stable, improved, or resolved \cite{gilman2012comparison,bannur2024maira2groundedradiologyreport}. However, many vision--language models for chest radiography are trained on static image--report pairs \cite{tiu2022expert,zhang2022contrastive,ko2025bringing,Bustos_2020}. Despite advances in thoracic finding recognition \cite{irvin2019chexpertlargechestradiograph,tiu2022expert} and radiology report generation \cite{chen2020r2gen,chen2021r2gencmn,bannur2024maira2groundedradiologyreport}, their ability to reason about localized changes between examinations remains insufficiently understood. Moreover, while longitudinal supervision is often derived from radiology reports, evaluating progression directly against expert judgments from paired images remains an important gap.

Temporally grounded reasoning requires connecting three dimensions: \emph{what} finding is being assessed, \emph{where} it occurs, and \emph{how} it evolves across studies. Existing datasets address these dimensions only partially. CheXpert \cite{chambon2024chexpertplusaugmentinglarge,irvin2019chexpertlargechestradiograph}, MIMIC-CXR \cite{johnson2019mimiccxrjpglargepubliclyavailable}, and ReXGradient \cite{zhang2025rexgradient160klargescalepubliclyavailable} provide large-scale image--report supervision, while MS-CXR-T \cite{bannur2023learningexploittemporalstructure} supplies paired-study progression labels and Chest ImaGenome \cite{wu2021chestimagenomedatasetclinical} provides anatomical grounding and report-derived comparison relations. Nevertheless, these resources do not jointly provide region-level progression labels, explicit cross-time pathology correspondence, and multi-source evaluation. As illustrated in Fig.~\ref{fig:chextemporal_annotations}, the same finding can worsen in one region while newly appearing in another, a distinction that a single image-level progression label cannot sufficiently capture.

We introduce \textbf{CheXTemporal}, a multi-source dataset for temporally grounded reasoning over paired prior--current CXRs. CheXTemporal comprises an expert-annotated gold dataset with 1,787 progression labels and 3,420 unique bounding boxes, alongside a large-scale silver dataset containing 282,214 automatically annotated examples. Spanning CheXpert, MIMIC-CXR, and ReXGradient, CheXTemporal provides five progression categories: \emph{new}, \emph{worse}, \emph{stable}, \emph{improved}, and \emph{resolved}, together with spatial annotations for localizable findings. Its gold annotations explicitly connect finding identity (\emph{what}), spatial localization (\emph{where}), and temporal progression (\emph{how}) through radiologist-drawn boxes and cross-time region correspondence. Table~\ref{tab:main_comparison} compares CheXTemporal with existing CXR datasets. We use CheXTemporal gold to benchmark existing vision--language CXR models on finding-level progression classification and spatial grounding, revealing limitations in fine-grained temporal reasoning and localization.

We further introduce \textbf{Prog-JEPA}, a progression-aware temporal reasoning framework built on the VL-JEPA embedding-prediction paradigm \cite{chen2026vljepa}. Prog-JEPA combines paired-image representations from BioViL-T with progression-adapted text embeddings and a finding-conditioned predictor trained on CheXTemporal silver annotations. We investigate alternative formulations of temporal supervision and evaluate their transfer ability to independent progression classification benchmarks and progression-specific spatial grounding.

Our contributions are summarized as follows:
\begin{itemize}
    \item We introduce \textbf{CheXTemporal}, a multi-source dataset for temporally grounded chest X-ray reasoning, with expert-annotated gold and large-scale silver subsets, finding-level spatial--temporal alignment, and five-class progression supervision.

    \item We introduce \textbf{Prog-JEPA}, a progression-aware joint-embedding predictive model trained on CheXTemporal silver annotations, and evaluate its transfer capability to independent temporal classification benchmarks and progression-specific spatial grounding.

    \item We establish a unified finding-level evaluation framework and systematically benchmark vision--language CXR models on progression classification, spatial grounding, and cross-source generalization.
\end{itemize}

\begin{table*}[t]
\centering
\caption{
Comparison of chest X-ray datasets.
CheXTemporal uniquely combines finding-level temporal
supervision, spatial--temporal alignment, a five-class
progression taxonomy, and multi-source coverage.
Gold and silver annotation counts refer to
progression labels.
}
\label{tab:main_comparison}

\scriptsize
\setlength{\tabcolsep}{5pt}
\renewcommand{\arraystretch}{1.15}

\begin{tabular}{@{}lcccccc@{}}
\toprule
\textbf{Property}
& \textbf{CheXpert}
& \textbf{MIMIC-CXR}
& \textbf{ReXGradient}
& \textbf{MS-CXR-T}
& \textbf{ImaGenome}
& \textbf{CheXTemporal} \\
\midrule

\multicolumn{7}{@{}l}{\textit{Dataset scale}} \\
\midrule

Patients
& 65,240
& 65,379
& 109,487
& 800
& 500 (gold)
& \textbf{197 / 34,296} \\

Gold annotations
& --
& --
& --
& 1,326
& 5,433
& \textbf{1,787} \\

Silver annotations
& --
& --
& --
& --
& --
& \textbf{282,214} \\

Findings
& --
& --
& --
& 5
& 55
& \textbf{11} \\

\midrule
\multicolumn{7}{@{}l}{\textit{Temporal supervision}} \\
\midrule

Progression classes
& --
& --
& --
& 3
& 3
& \textbf{5} \\

Finding-level temporal
& $\times$
& $\times$
& $\times$
& \checkmark
& \checkmark
& \checkmark \\

\midrule
\multicolumn{7}{@{}l}{\textit{Spatial grounding and coverage}} \\
\midrule

Spatial supervision
& $\times$
& $\times$
& $\times$
& $\times$
& \checkmark
& \checkmark \\

\textbf{Spatial--temporal alignment}
& $\times$
& $\times$
& $\times$
& $\times$
& $\times$
& \textbf{\checkmark} \\

\textbf{Multi-source}
& $\times$
& $\times$
& $\times$
& $\times$
& $\times$
& \textbf{\checkmark} \\

\bottomrule
\end{tabular}

\vspace{3pt}
\begin{minipage}{\textwidth}
\footnotesize
\textit{Notes:}
CheXTemporal patient counts are gold / silver.
MS-CXR-T and Chest ImaGenome are curated subsets
of MIMIC-CXR.
Chest ImaGenome additionally provides 217,013 studies
and 242,072 images, but these are not directly
comparable to pair--finding annotation counts.
Spatial--temporal alignment denotes explicitly aligned
finding-level localization and progression annotations
across paired studies.

Dataset references:
CheXpert~\cite{chambon2024chexpertplusaugmentinglarge,irvin2019chexpertlargechestradiograph};
MIMIC-CXR~\cite{johnson2019mimiccxrjpglargepubliclyavailable};
ReXGradient~\cite{zhang2025rexgradient160klargescalepubliclyavailable};
MS-CXR-T~\cite{bannur2023learningexploittemporalstructure};
Chest ImaGenome~\cite{wu2021chestimagenomedatasetclinical}.
\end{minipage}
\end{table*}

\section{Related Work}

\noindent\textbf{Chest radiography datasets and temporal benchmarks.}
Large-scale CXR datasets, including CheXpert~\cite{irvin2019chexpertlargechestradiograph,chambon2024chexpertplusaugmentinglarge}, MIMIC-CXR~\cite{johnson2019mimiccxrjpglargepubliclyavailable}, and ReXGradient~\cite{zhang2025rexgradient160klargescalepubliclyavailable}, provide image--report supervision but limited explicit annotations of localized temporal change. Chest ImaGenome~\cite{wu2021chestimagenomedatasetclinical} provides anatomical scene graphs and comparison relations, while MS-CXR-T~\cite{bannur2023learningexploittemporalstructure} supplies paired-study progression labels. LUNGUAGE~\cite{moon2026lunguage} addresses structured longitudinal report interpretation, and MI-CXR~\cite{cho2026micxr} evaluates multi-interval visual reasoning. PadChest-GR~\cite{castro2025padchestgr} combines radiologist-drawn boxes with report-derived progression labels, but its progression-annotated subset is smaller and boxes cover only the current examination, without explicit cross-time pathology correspondence. CheXTemporal instead provides region-specific expert progression labels, matched pathology boxes across studies, and multi-source silver supervision.

\noindent\textbf{Vision--language models for chest radiography.}
Medical vision--language representation learning includes BioViL~\cite{Boecking_2022}, GLoRIA~\cite{huang2021gloria}, MedCLIP~\cite{wang2022medclipcontrastivelearningunpaired}, ConVIRT~\cite{zhang2022contrastive}, and BiomedCLIP~\cite{zhang2025biomedclipmultimodalbiomedicalfoundation}; CheXagent~\cite{chen2024chexagent} extends CXR interpretation through instruction tuning. Longitudinal approaches such as BioViL-T~\cite{bannur2023learningexploittemporalstructure}, TILA~\cite{ko2026temporalinversionlearninginterval}, TempA-VLP~\cite{Yang2025TempAVLPTV}, and ALTA~\cite{Lian_2025} incorporate temporal information through paired-image representations or temporal learning objectives. BAAP~\cite{chen2026baap} introduces anatomy-aware progression prediction, while MAIRA-2~\cite{bannur2024maira2groundedradiologyreport} uses prior imaging and reports for grounded report generation. These methods do not directly establish progression performance on expert-drawn pathology regions matched across examinations. VL-JEPA~\cite{chen2026vljepa} predicts text embeddings instead of generating tokens; Prog-JEPA adapts this paradigm to finding-conditioned temporal CXR reasoning.

\section{Dataset}

\begin{figure}[t]
  \centering
  \includegraphics[width=\linewidth]{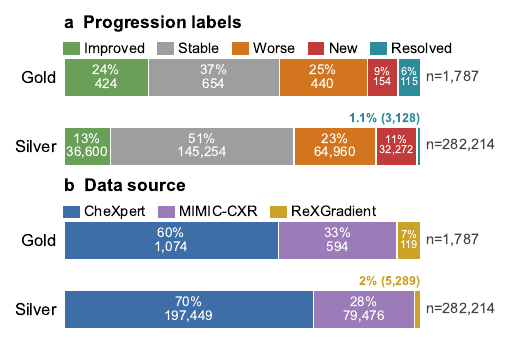}
  \caption{Composition of CheXTemporal; gold counts distinguish progression labels from pair--finding examples.
  (a) Progression-label distribution. (b) Data source.
  Gold: 197 patients, 1,787 progression labels across 1,497 pair--finding examples, 3,420 manual bounding boxes.
  Silver: 34,296 patients, 282,214 examples, 168,140 automatic segmentation masks,
  282,214 automatic anatomy labels, and 278,953 static and 416,976 dynamic report sentences.}
  \label{fig:dataset_stats}
\end{figure}

We construct two complementary datasets for temporally grounded reasoning in chest radiography: a radiologist annotated gold dataset and a large--scale automatically derived silver dataset. The gold dataset provides expert finding--level spatial and temporal annotations for precise evaluation, while the silver dataset scales the same supervision structure across substantially more study pairs and clinical sources. Dataset statistics are summarized in Fig.~\ref{fig:dataset_stats}. The annotation structure captures three complementary
dimensions of longitudinal reasoning: \emph{what}
finding is being assessed, represented by the target
pathology; \emph{where} it occurs, represented by
spatial annotations; and \emph{how} it evolves,
represented by the five-class progression taxonomy.
Together, these annotations establish finding-level
spatial--temporal relationships across paired studies.

\subsection{Annotation Schema}
\label{sec:annotation_schema}

Each \emph{study pair} consists of a prior and a current frontal CXR from one patient. A \emph{pair--finding example} associates this pair with one target finding; a pair can therefore contribute multiple examples. In gold, a finding may occupy multiple \emph{regions}, each assigned a progression label and a cross-time correspondence identity. A \emph{bounding box} localizes that region on the prior or current image when visible; the same region can have a box at both time points. Thus, the counts of pairs, pair--finding examples, progression labels, regions, and boxes are not interchangeable.

For \emph{what}, the target is one of eleven thoracic findings; for \emph{where}, gold provides pathology boxes and cross-time correspondence while silver supplies report-derived anatomical labels and anatomy masks; for \emph{how}, both use the five progression categories \emph{new}, \emph{worse}, \emph{stable}, \emph{improved}, and \emph{resolved}. New and Resolved indicate appearance and disappearance, respectively; Worse and Improved indicate increasing and decreasing severity; Stable denotes no meaningful interval change. Figure~\ref{fig:chextemporal_annotations} illustrates the two annotation schemes, and Appendix Table~\ref{tab:annotation_schema} summarizes their granularity; further dataset and annotation details appear in Appendices~A--C.

\subsection{Gold Dataset}

\noindent\textbf{Data and finding set.}
The gold dataset comprises paired prior--current frontal CXRs from CheXpert~\cite{irvin2019chexpertlargechestradiograph,chambon2024chexpertplusaugmentinglarge}, MIMIC-CXR~\cite{johnson2019mimiccxrjpglargepubliclyavailable}, and ReXGradient~\cite{zhang2025rexgradient160klargescalepubliclyavailable}. Each current study is paired with the most recent preceding frontal study available for the same patient. Target findings are identified using report context and standardized with CheXbert~\cite{smit2020chexbertcombiningautomaticlabelers}. We consider eleven findings: Atelectasis, Cardiomegaly, Consolidation, Edema, Enlarged Cardiomediastinum, Lung Lesion, Lung Opacity, Pleural Effusion, Pleural Other, Pneumonia, and Pneumothorax. We exclude No Finding (not a pathology), Support Device (primarily positional changes), and Fracture (less naturally captured by the five-class progression taxonomy). Radiologists assess paired images without access to reports and assign \emph{new}, \emph{worse}, \emph{stable}, \emph{improved}, or \emph{resolved} labels.

\noindent\textbf{Annotation protocol.}
Annotations were collected using a Gradio-based interface~\cite{abid2019gradio} displaying paired radiographs side by side (Appendix Fig.~\ref{fig:gradio_platform}). Radiologists assess one target finding at a time, drawing boxes for up to five regions per image to capture multifocal disease. Correspondence identities link matched regions across studies through visual assessment; no registration or automated matching is used. Images are resized to a 1024-pixel shorter side while preserving aspect ratio. The workflow and region limit were developed with annotator input; full annotation-platform instructions are provided in Appendix~E.

\noindent\textbf{Annotators.}
Seven physicians with radiology training, including academic faculty, board-certified attending radiologists, and a physician with residency-level radiology training, provided primary annotations, with one primary annotation per gold example. An additional board-certified thoracic radiologist participated in the inter-rater assessment, bringing the total to eight. For this assessment, radiologists independently re-annotated cases they had not previously annotated, blinded to the original labels. Annotators estimated approximately 1--2 minutes per pair--finding annotation; times were not formally recorded.

\noindent\textbf{Spatial and temporal supervision.}
Gold annotations provide approximate pathology boxes rather than segmentation masks. Edema, due to its diffuse presentation, receives progression labels without localization. For other findings, radiologists drew 3,420 boxes (1,679 prior; 1,741 current) across 1,824 regions. Of these, 1,568 (86\%) are marked at both time points, establishing explicit correspondence. Current-only regions are predominantly \emph{new} (91\% of 158), while prior-only regions are predominantly \emph{resolved} (81\% of 98). Multifocal disease is common: 461 pair--finding examples contain at least two regions, and 251 exhibit different progression trajectories across regions. Overall, gold contains 1,787 progression labels across 1,497 pair--finding examples (Fig.~\ref{fig:dataset_stats}).

\noindent\textbf{Quality control and inter-radiologist agreement.}
Annotations link study pairs, findings, boxes, correspondence identities, and progression labels using standardized instructions and label definitions. A second radiologist independently re-annotated cases without access to the original labels. Across 415 matched cases, unweighted five-class Cohen's $\kappa$ was 0.339 on 305 single-label cases, with 54.8\% exact agreement; quadratic-weighted $\kappa$ was 0.504, and 78\% of disagreements concerned Stable versus a change label. Across 497 matched spatial cases, mean best-match IoU was 0.431 (prior) and 0.427 (current). These findings identify subtle change versus stability as a major source of ambiguity and provide a human reference for model evaluation. Full analyses are in Appendix~\ref{sec:supp_gold_agreement}.

\subsection{Silver Dataset}

\noindent\textbf{Motivation and scope.}
To complement the manually curated gold dataset, we construct a large--scale silver dataset with automatically derived temporal and spatial supervision. As shown in Fig.~\ref{fig:dataset_stats}, the silver dataset contains 282,214 pair--finding examples spanning CheXpert, MIMIC-CXR, and ReXGradient.

\noindent\textbf{Pair construction.}
Temporal pairs are constructed by linking each current study to a corresponding prior study for the same patient. We restrict studies to frontal views to ensure consistent evaluation across sources.

\noindent\textbf{Silver label generation.}
Silver annotations are generated using MedGemma-27B \cite{sellergren2026medgemmatechnicalreport} with structured prompts that ask the model to compare the current and prior reports as an expert radiologist. Given the findings and impression sections from both reports, the model produces two forms of supervision. First, it assigns a finding--level progression label from the same five--class taxonomy used in the gold dataset: \emph{new}, \emph{worse}, \emph{stable}, \emph{improved}, or \emph{resolved}. Second, it labels report sentences as \emph{static} if they describe the current study alone or \emph{dynamic} if they express temporal comparison or change. Outputs are generated deterministically and post--processed to enforce closed vocabularies for findings, progression labels, and anatomical regions. Full prompting details and templates are provided in Appendix~D (Figs.~\ref{fig:findings_system_prompt} and~\ref{fig:sentence_system_prompt}).

\noindent\textbf{Anatomical grounding.}
To provide scalable spatial supervision, we associate each silver finding with anatomical regions derived from report text and segmentation masks. Anatomical mentions produced by MedGemma are mapped to a controlled vocabulary of chest regions. We then use the CXAS segmentation model~\cite{seibold2023accuratefinegrainedsegmentationhuman} to obtain anatomy masks and retain only regions corresponding to the target finding. Only anatomically consistent segmentation masks are retained as silver spatial annotations; cases with missing required anatomy classes, implausible anatomy coverage, or unreliable anatomy mappings are excluded. Datasets exhibiting systematic segmentation failures due to domain mismatch are excluded from the retained mask subset. These masks identify anatomical regions rather than delineating pathology boundaries; examples and filtering details are given in the Appendix (Fig.~\ref{fig:silver_segmentation_masks}).

\noindent\textbf{Post-processing and filtering.}
All generated records are deterministically audited prior to evaluation. Outputs with malformed JSON or progression labels outside the five-class taxonomy are discarded. Anatomical phrases outside the closed vocabulary are normalized through manually verified rule-based mappings into a controlled 30-region anatomical label space. Less than 1\% of generated outputs were discarded due to malformed structure or invalid progression labels, while approximately 7\% required anatomy normalization through rule-based mapping. The complete auditing and normalization pipeline, including representative anatomy mappings, is described in the Appendix (Fig.~\ref{fig:audit-mapping}).

\noindent\textbf{Validation of silver supervision.}
We manually reviewed silver annotations for 55 randomly sampled paired-report cases from each data source to assess fidelity to the underlying radiology reports, rather than independent clinical correctness. The cases were divided among the first three authors, with one reviewer per case. Two participating radiologists advised on the prompting and anatomy definitions and performed limited checks during prompt development; the validation sample was not independently re-read by expert radiologists, and masks were not manually validated case by case. Progression and anatomy labels were correct for every reviewed finding, while static/dynamic sentence-label accuracy ranged from 95.2\% to 100\% (Appendix Fig.~\ref{fig:manual}).

To further assess cross-model consistency, we compared MedGemma silver annotations with independently generated Qwen3-32B-AWQ~\cite{qwen3technicalreport} annotations on overlapping examples. Across 281,836 matched findings, progression-label agreement was 80.1\% (Cohen's $\kappa = 0.71$, substantial agreement). Most disagreements were not contradictory: 73.8\% were between Stable and a change label, typically reflecting differing thresholds for what constitutes interval change. The pattern of cross-model disagreement on silver resembles inter-radiologist disagreement on gold: most disagreements concern whether a finding is stable or has changed (73.8\% on silver and 78\% on gold), rather than the direction of change. Because these comparisons involve different cases and annotation sources, their similar patterns suggest that distinguishing subtle change from stability is a recurring source of ambiguity, rather than establishing direct gold--silver label agreement. Static/dynamic sentence classification achieved 90.3\% agreement ($\kappa = 0.80$) across 630,540 sentences. Fine-grained anatomical agreement was lower, reflecting genuine ambiguity in how reports specify sub-regional location; most of these disagreements (65\%) are nonetheless resolved at the laterality level. These findings demonstrate substantial cross-model consistency for progression and sentence-level supervision, while identifying fine-grained anatomical localization as a less reproducible component. Detailed cross-model agreement statistics are provided in Appendix~\ref{sec:supp_silver_crossmodel}.

\subsection{Dataset Splits and Counting Units}
\label{sec:dataset_splits}

For silver-supervised training, we use a fixed, pair-level 90/10 training/validation split (seed 42). The training partition contains 246,454 pair--finding examples: 32,221 Improved, 128,123 Stable, 56,842 Worse, 26,597 New, and 2,671 Resolved. The silver partition is used for model training and validation, while the radiologist-annotated gold dataset provides the expert evaluation benchmark. External evaluation pairs sharing a patient, study, or image with silver training data are excluded, and conflicting or duplicate labels are discarded. In the reported cohort summaries, mean ages are 64.4 years (gold) and 63.5 years (silver), with 52.8\% and 55.3\% male patients; median prior--current intervals are 13 and 7 days, respectively, where dates are available. Source-level comparisons and metadata limitations are given in Appendix Table~\ref{tab:supp_dataset_characteristics}.

Different evaluation protocols operate on different units of the same gold annotations. The gold dataset contains 1,787 progression labels across 1,497 pair--finding groups. The progression-grounding subset includes 1,565 pair--finding--label annotations; excluding 76 without a current-study box leaves 1,489 current-study annotations across 673 study pairs. Appendix Table~\ref{tab:gold_counting_units} reconciles these counting units.

Ethical considerations are discussed in Appendix~\ref{sec:supp_ethics}.
\section{Benchmark}
\label{sec:benchmark}

CheXTemporal evaluates two aspects of finding-conditioned longitudinal interpretation: \emph{how} a named finding changes (progression classification) and \emph{where} it is supported in the images (temporally grounded localization). Both tasks condition on \emph{what} finding is assessed. This section specifies the evaluation inputs, outputs, datasets, and metrics independently of any particular model.

\subsection{Progression Classification}
\label{sec:benchmark_classification}

Given a prior CXR $x_p$, current CXR $x_c$, and target finding $f$, a model assigns progression scores to $\mathcal{C}=\{\mathrm{improved},\mathrm{stable},\mathrm{worse},\mathrm{new},\mathrm{resolved}\}$. On CheXTemporal gold, evaluation uses 1,497 pair--finding groups: 1,239 have a single reference progression class and 258 have multiple classes owing to different regional trajectories. The single-label groups comprise 275 Improved, 509 Stable, 286 Worse, 97 New, and 72 Resolved examples. For single-label groups, the prediction is the five-class argmax; for a multi-label group with $k$ reference labels, the prediction is the set of top-$k$ classes. The combined score is mean label-set Jaccard, $|\hat{Y}\cap Y|/|\hat{Y}\cup Y|$, across all groups. We additionally report five-class macro F1, per-class F1, and single-label accuracy on the 1,239 single-label groups. Because the task concerns \emph{how} a specified finding changes, we foreground metrics over progression labels rather than disease-category aggregates; per-class F1 makes limitations on individual temporal states visible.

We also evaluate three-class progression transfer on MS-CXR-T~\cite{bannur2023learningexploittemporalstructure} (339 pairs: 51 Improved, 124 Stable, 164 Worse) and Chest ImaGenome~\cite{wu2021chestimagenomedatasetclinical} (348 temporal pairs: 50 Improved, 184 Stable, 114 Worse). For these benchmarks, predictions are restricted to Improved, Stable, and Worse, and we report macro F1, per-class F1, and accuracy. Paired 95\% confidence intervals for classification comparisons use 2,000 item-level bootstrap resamples.

\subsection{Temporally Grounded Localization}
\label{sec:benchmark_grounding}

Given an image pair, a target finding, and a text direction describing either the finding or its progression, the output is a spatial activation map on the current CXR. Evaluation uses the union of the radiologist-drawn current-study boxes as the reference mask. The gold grounding subset contains 1,565 pair--finding--label annotations; excluding 76 without a current-study box leaves 1,489 annotations from 673 pairs (354 Improved, 560 Stable, 381 Worse, 142 New, 52 Resolved; 867 CheXpert, 519 MIMIC-CXR, 103 ReXGradient; Sec.~\ref{sec:dataset_splits}). For every map, we report pixel AUROC, pointing-game accuracy (whether the maximum activation falls inside the boxes), area-matched mean intersection over union (mIoU), and contrast-to-noise ratio (CNR), which measures inside/outside activation separation. Significance is assessed using 2,000 paired bootstrap resamples over image pairs.

We evaluate two settings: \emph{finding-presence grounding}, which localizes a named finding using present-versus-absent text contrasts, and \emph{progression-specific grounding}, which localizes progression direction while controlling for specified finding-presence directions. For the latter, we contrast the target progression class against the other four classes and orthogonally remove three finding-presence contrasts. We additionally evaluate full class-sentence, unprojected progression-contrast, and finding-presence directions, including four-paraphrase variants. Spatial maps are derived from patch--text similarities. This projection removes the specified presence directions, not necessarily every possible presence cue. Mathematical definitions and map-construction details are provided in Appendix~\ref{sec:supp_grounding}.

\section{Prog-JEPA}
\label{sec:prog_jepa}

Prog-JEPA adapts the VL-JEPA embedding-prediction paradigm~\cite{chen2026vljepa} to finding-conditioned temporal CXR reasoning. Rather than predicting only classification logits, it learns to map paired radiographs into a progression-aware text-embedding space, allowing us to assess whether silver supervision captures finding-specific temporal semantics.

\noindent\textbf{Problem formulation.}
Given a prior radiograph $x_p$, current radiograph $x_c$, and finding $f$, the task is to predict $y\in\mathcal{C}=\{\mathrm{improved},\mathrm{stable},\mathrm{worse},\mathrm{new},\mathrm{resolved}\}$. Prog-JEPA maps the image pair and finding query to a normalized text-space embedding $\hat{s}$ and retrieves the closest progression-class sentence at inference. The training and inference framework is illustrated in Appendix Fig.~\ref{fig:model_architecture}.

\noindent\textbf{Architecture.}
Prog-JEPA consists of three components: (1) a BioViL-T temporal image encoder~\cite{bannur2023learningexploittemporalstructure} that represents current-image features and prior-conditioned differences as global and patch embeddings; (2) a progression-adapted CXR-BERT text encoder \cite{bannur2023learningexploittemporalstructure} that maps the sentence ``$f$ is $k$.'' to a normalized target embedding $s_k$ for class $k$; and (3) a transformer predictor initialized from Llama-3.2-1B~\cite{meta2024llama32modelcard}, which combines the temporal image features with the query ``What is the progression of $f$?'' to predict $\hat{s}$. Both target and predicted embeddings are 128-dimensional. The adapted text encoder remains frozen during Prog-JEPA training. Full architectural details are provided in Appendix~\ref{sec:supp_prog_jepa_architecture}.

\noindent\textbf{Progression-aware text adapter.}
Because the original CXR-BERT embeddings poorly distinguish progression descriptions of the same finding, we train a LoRA adapter on 220 templated sentences spanning 11 findings, five progression classes, and four paraphrases per class. A supervised contrastive objective ($\tau=0.07$) treats sentences as positives only when both finding and progression class match. The learned embedding organization is visualized in Appendix~J (Fig.~\ref{fig:text_embeddings}).

\noindent\textbf{Objective and inference.}
For each silver training tuple $(x_p,x_c,f,y)$, we minimize class-weighted InfoNCE over the five progression-class sentences for the same finding:
\begin{equation}
\mathcal{L}=-w_y\log\frac{\exp(\cos(\hat{s},s_y)/\tau)}{\sum_{k\in\mathcal{C}}\exp(\cos(\hat{s},s_k)/\tau)},\quad\tau=0.07.
\end{equation}
Class weights follow the effective-number scheme ($\beta=0.99999$) and sum to five. At inference, the predicted class is $\arg\max_k[\cos(\hat{s},s_k)+b_k]$, with class-score offsets selected on silver validation (Appendix~\ref{sec:supp_training_implementation}).

\noindent\textbf{Alternative supervision.}
We evaluate two variants using the same temporal encoder and finding-conditioned predictor. \emph{Dynamic Prog-JEPA} replaces fixed progression-class targets with dynamically labeled report sentences and uses in-batch contrastive learning. \emph{Laterality Prog-JEPA} augments progression targets with left, right, or bilateral anatomy when the silver annotation models agree; laterality is omitted from the query and marginalized at inference. Implementation differences, including text encoders and effective batch sizes, are detailed in Appendix~\ref{sec:supp_training_implementation}; results are discussed in Sec.~\ref{sec:results_ablations}.

\section{Results}
\label{sec:results}

We evaluate pretrained models' temporal reasoning capabilities, the benefits of large-scale silver supervision, and the spatial grounding of progression predictions.

\subsection{Experimental Setup}
\label{sec:results_setup}

\noindent\textbf{Pretrained models and zero-shot protocol.}
We prioritize publicly released CXR vision--language models with available pretrained checkpoints and consider documented training data to reduce potential evaluation contamination. The five pretrained baselines are BioViL~\cite{Boecking_2022}, BiomedCLIP~\cite{zhang2025biomedclipmultimodalbiomedicalfoundation}, BioViL-T~\cite{bannur2023learningexploittemporalstructure}, TILA~\cite{ko2026temporalinversionlearninginterval}, and ALTA~\cite{Lian_2025}. BioViL and BiomedCLIP primarily use static image--text pretraining, whereas the other models incorporate temporal information. We use 37 progression phrases for zero-shot classification, with full scoring details provided in Appendix~F. The compared pretrained models are evaluated without task-specific fine-tuning.

\noindent\textbf{Supervised baseline.}
A deliberately simple supervised BioViL-T baseline uses unfrozen image and text encoders with a linear five-logit classifier over concatenated pair and finding embeddings, trained on the same silver examples, split, and augmentations as Prog-JEPA. It tests the utility of silver data under conventional classification; it is not a capacity-matched ablation of the Prog-JEPA predictor.

For Prog-JEPA and its alternative supervision variants, training configurations, image-encoder fine-tuning, text adapters, class-score calibration, and implementation differences are detailed in Appendix~\ref{sec:supp_training_implementation}.

\noindent\textbf{Grounding comparisons.}
For grounding, we compare the original BioViL-T image/text encoders with the fine-tuned Prog-JEPA image encoder and adapted text encoder. We also examine the adapted text encoder paired with the frozen BioViL-T image encoder, and the other pretrained baselines; model-specific map constructions appear in Appendix~\ref{sec:supp_grounding}. Patch/text similarity maps are used instead of predictor attribution because the predictor's all-to-all attention does not preserve spatial token correspondence.

\subsection{Zero-Shot Benchmark}
\label{sec:results_zeroshot}

Existing pretrained CXR representations struggle with fine-grained progression classification, particularly with identifying temporal stability. On CheXTemporal gold, BioViL-T~\cite{bannur2023learningexploittemporalstructure} is the strongest of the five zero-shot baselines (Table~\ref{tab:vljepa_gold}), yet its Stable F1 is substantially below its Worse F1. Across models, Stable is poorly recognized, even when salient worsening can be detected. The first block of Table~\ref{tab:vljepa_gold} reports these pretrained baselines; its remaining rows compare methods trained on silver, discussed in Sec.~\ref{sec:learned_results}.

\begin{table}[t]
\centering
\caption{Progression classification on CheXTemporal gold (1,497 groups). Combined is mean label-set Jaccard; F1 uses 1,239 single-label groups. FT denotes image-encoder fine-tuning.}
\label{tab:vljepa_gold}
\vspace{-4pt}
\setlength{\tabcolsep}{3pt}
\resizebox{\columnwidth}{!}{%
\begin{tabular}{@{}lccccccc@{}}
\toprule
Model & Combined & Macro F1 & Impr. & Stable & Wors. & New & Resolved \\
\midrule
BioViL~\cite{Boecking_2022} & .243 & .182 & .137 & .062 & .430 & .146 & .133 \\
BiomedCLIP~\cite{zhang2025biomedclipmultimodalbiomedicalfoundation} & .223 & .150 & .115 & .104 & .354 & .060 & .117 \\
BioViL-T~\cite{bannur2023learningexploittemporalstructure} & .307 & .249 & .297 & .145 & .417 & .212 & .175 \\
TILA~\cite{ko2026temporalinversionlearninginterval} & .261 & .208 & .230 & .107 & .373 & .202 & .129 \\
ALTA~\cite{Lian_2025} & .256 & .213 & .318 & .132 & .278 & .176 & .159 \\
\midrule
Supervised & .457 & .393 & .463 & .515 & .488 & .272 & .228 \\
Prog-JEPA (5-way, FT) & \textbf{.480} & .394 & .370 & \textbf{.612} & .436 & \textbf{.308} & .244 \\
Dynamic Prog-JEPA (FT) & .331 & .276 & .394 & .205 & .419 & .200 & .164 \\
Laterality Prog-JEPA (FT) & .459 & .385 & .403 & .577 & .414 & .287 & .243 \\
\bottomrule
\end{tabular}%
}
\vspace{-6pt}
\end{table}

\subsection{Learning from Silver Supervision}
\label{sec:learned_results}

Silver-supervised Prog-JEPA improves progression classification on CheXTemporal gold and generalizes to both external benchmarks. Compared with a conventional supervised classifier trained on the same silver examples, Prog-JEPA achieves a higher combined score on gold and higher macro F1 on MS-CXR-T and Chest ImaGenome. However, improvements vary across progression categories, and gold macro F1 remains comparable to the supervised baseline. Tables~\ref{tab:vljepa_gold}, \ref{tab:vljepa_mscxrt}, and~\ref{tab:vljepa_imagenome} report the image-encoder fine-tuned configuration of each Prog-JEPA supervision variant for consistent comparison. Complete results, including frozen and unshifted configurations, appear in Appendix~\ref{sec:supp_progression_tables}.

\begin{table}[t]
\centering
\caption{Progression classification on MS-CXR-T (339 pairs). All scores are F1 except accuracy. FT denotes image-encoder fine-tuning.}
\label{tab:vljepa_mscxrt}
\vspace{-4pt}
\setlength{\tabcolsep}{3pt}
\resizebox{\columnwidth}{!}{%
\begin{tabular}{@{}lccccc@{}}
\toprule
Model & Macro F1 & Impr. & Stable & Wors. & Acc. \\
\midrule
BioViL~\cite{Boecking_2022} & .381 & .374 & .137 & .632 & .460 \\
BiomedCLIP~\cite{zhang2025biomedclipmultimodalbiomedicalfoundation} & .365 & .279 & .184 & .631 & .460 \\
BioViL-T~\cite{bannur2023learningexploittemporalstructure} & .372 & .278 & .177 & .660 & .487 \\
TILA~\cite{ko2026temporalinversionlearninginterval} & .420 & .359 & .248 & .654 & .490 \\
ALTA~\cite{Lian_2025} & .331 & .321 & .161 & .510 & .366 \\
\midrule
Supervised & .420 & .433 & .120 & .707 & .513 \\
Prog-JEPA (5-way, FT) & \textbf{.512} & .402 & .410 & \textbf{.724} & \textbf{.549} \\
Dynamic Prog-JEPA (FT) & .471 & .432 & .278 & .702 & .534 \\
Laterality Prog-JEPA (FT) & .481 & .422 & .350 & .671 & .507 \\
\bottomrule
\end{tabular}%
}
\vspace{-6pt}
\end{table}

\begin{table}[t]
\centering
\caption{Progression classification on Chest ImaGenome (348 pairs). All scores are F1 except accuracy. FT denotes image-encoder fine-tuning.}
\label{tab:vljepa_imagenome}
\vspace{-4pt}
\setlength{\tabcolsep}{3pt}
\resizebox{\columnwidth}{!}{%
\begin{tabular}{@{}lccccc@{}}
\toprule
Model & Macro F1 & Impr. & Stable & Wors. & Acc. \\
\midrule
BioViL~\cite{Boecking_2022} & .322 & .297 & .155 & .513 & .351 \\
BiomedCLIP~\cite{zhang2025biomedclipmultimodalbiomedicalfoundation} & .276 & .140 & .179 & .511 & .351 \\
BioViL-T~\cite{bannur2023learningexploittemporalstructure} & .372 & .362 & .225 & .528 & .405 \\
TILA~\cite{ko2026temporalinversionlearninginterval} & .336 & .293 & .257 & .458 & .353 \\
ALTA~\cite{Lian_2025} & .266 & .276 & .011 & .510 & .273 \\
\midrule
Supervised & .509 & .388 & .521 & .619 & .526 \\
Prog-JEPA (5-way, FT) & \textbf{.586} & .403 & .695 & \textbf{.659} & \textbf{.632} \\
Dynamic Prog-JEPA (FT) & .333 & .384 & .063 & .552 & .382 \\
Laterality Prog-JEPA (FT) & .547 & .387 & \textbf{.697} & .557 & .598 \\
\bottomrule
\end{tabular}%
}
\vspace{-6pt}
\end{table}

\noindent\textbf{CheXTemporal gold.}
Fine-tuned Prog-JEPA attains the highest combined score (0.480) and single-label accuracy (0.481). Its combined score exceeds the supervised baseline by 0.023 (95\% CI: 0.001--0.046) and single-label accuracy by 0.032 (0.005--0.059), while five-way macro F1 is effectively tied (0.394 versus 0.393). Prog-JEPA is stronger on Stable (+0.097 F1), New, and Resolved, while the supervised baseline is stronger on Improved (+0.094) and Worse (+0.052). Fine-tuned Prog-JEPA leads on CheXpert (combined score +0.031, 95\% CI: 0.001--0.061) and is tied on MIMIC; ReXGradient has only 85 groups. All five zero-shot models trail fine-tuned Prog-JEPA significantly in combined score (by 0.17--0.26), macro F1, and Stable F1, with systematic underprediction of Stable (e.g., BioViL-T predicts Stable for 111 groups versus 509 Stable references).

\noindent\textbf{External benchmarks.}
Fine-tuned five-way Prog-JEPA achieves the highest macro F1 on both external benchmarks (Tables~\ref{tab:vljepa_mscxrt} and~\ref{tab:vljepa_imagenome}). It outperforms the supervised model trained on identical silver data by 0.092 on MS-CXR-T (95\% CI: 0.038--0.144) and 0.076 on Chest ImaGenome (0.027--0.127). It also outperforms all evaluated zero-shot representations, with macro-F1 margins of 0.09--0.18 on MS-CXR-T and 0.21--0.32 on Chest ImaGenome. A major difference is recognition of Stable: the supervised baseline achieves Stable F1 of 0.120 on MS-CXR-T, whereas fine-tuned Prog-JEPA reaches 0.410.

Overall, training on silver supervision improves generalization, especially on external three-class benchmarks and recognition of Stable. The modest gold gain and class-specific trade-offs keep this result more specific than a uniform improvement across all progression categories.

\subsection{Progression Grounding}
\label{sec:vljepa_grounding}

\begin{figure}[t]
  \centering
  \includegraphics[width=0.7\linewidth]{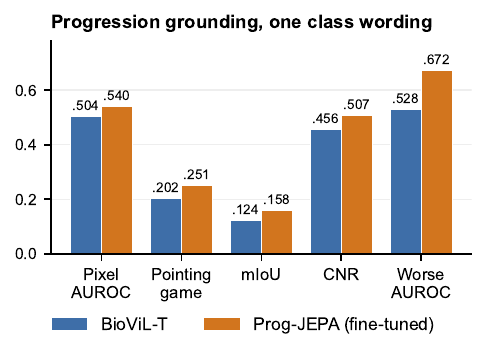}
  \vspace{-10pt}
  \caption{Progression-specific grounding on CheXTemporal gold (1,489 annotations). BioViL-T versus fine-tuned Prog-JEPA using one class wording; Worse AUROC is computed on worsening annotations.}
  \label{fig:grounding}
\end{figure}

We evaluate progression-specific localization using the protocol in Sec.~\ref{sec:benchmark_grounding}. Figure~\ref{fig:grounding} compares BioViL-T with fine-tuned Prog-JEPA under the primary single-wording setting; additional prompt, finding-presence, and baseline results are provided in Appendix~\ref{sec:supp_grounding}. Spatial maps use patch--text similarities rather than attribution through the Llama predictor, whose all-to-all attention does not preserve spatial token correspondence.

\noindent\textbf{Progression-specific grounding.}
After projecting out specified finding-presence directions, fine-tuned Prog-JEPA improves pixel AUROC from 0.504 to 0.540 (95\% CI for the difference: 0.024--0.046), pointing-game accuracy from 0.202 to 0.251, mIoU from 0.124 to 0.158, and CNR from 0.456 to 0.507 (Fig.~\ref{fig:grounding}). All four improvements are statistically significant under paired bootstrap evaluation. The largest class-specific gain is for Worse (AUROC 0.528 to 0.672). Improved, Stable, and New remain near chance, while both models score below chance for Resolved, whose pathology is absent from the current image (Appendix Table~\ref{tab:supp_ground_class}). Most improvement comes from the adapted text encoder, with additional benefit from image fine-tuning. Overall AUROC remains modest, and projection controls only the specified presence directions, not all possible presence cues.

\noindent\textbf{Additional grounding comparisons.}
Four-paraphrase directions yield consistent improvements, but finding-presence localization shows trade-offs: fine-tuned Prog-JEPA largely preserves BioViL-T's presence AUROC while reducing pointing-game accuracy. Among additional baselines, TILA~\cite{ko2026temporalinversionlearninginterval} achieves higher overall progression AUROC (0.565 vs.\ 0.540) but lower finding-presence AUROC (0.634 vs.\ 0.755). Prog-JEPA is strongest on Worse across evaluated metrics under the single-wording direction. Results for BioViL, BiomedCLIP, ALTA, alternative text encoders, and predictor attribution are provided in Appendix~\ref{sec:supp_grounding} and Fig.~\ref{fig:supp_grounding_variants}. These findings show improved progression-specific spatial evidence, particularly for worsening disease, without uniformly improved finding-presence localization.

\subsection{Ablations and Alternative Supervision}
\label{sec:results_ablations}

Fixed five-class progression supervision provides the most consistent overall performance across gold and external benchmarks (Tables~\ref{tab:vljepa_gold}, \ref{tab:vljepa_mscxrt}, and~\ref{tab:vljepa_imagenome}). Dynamic-sentence targets perform above zero-shot baselines on gold but below five-way Prog-JEPA; they transfer competitively to MS-CXR-T but poorly to Chest ImaGenome, largely due to underprediction of Stable. Laterality targets improve selected classes without consistently improving overall classification, with Worse remaining a weakness.

On gold, removing the silver-selected Resolved-score offset improves laterality performance, including Improved F1 after image fine-tuning, indicating a calibration trade-off. The offset does not affect external three-class evaluation. Detailed frozen and fine-tuned results, bootstrap intervals, and laterality readouts are provided in Appendix~\ref{sec:supp_progression_tables}. Because supervision variants also differ in text encoders and effective batch sizes, these comparisons do not isolate the effect of supervision targets alone (Appendix~\ref{sec:supp_training_implementation}).

\section{Conclusion}

We introduced CheXTemporal, a multi-source dataset for temporally grounded chest radiography, combining radiologist-annotated progression labels, pathology boxes, and cross-time region correspondence with large-scale silver supervision. It supports evaluation of five-class progression classification and spatial grounding over paired CXRs. Our benchmark exposes limitations in existing vision--language models, while silver-trained Prog-JEPA improves gold progression classification and outperforms a supervised baseline on MS-CXR-T and Chest ImaGenome. It also strengthens progression-specific grounding, particularly for worsening findings, although localized temporal reasoning remains challenging. The CheXTemporal dataset will be released publicly.

\section{LLM Usage Statement}
OpenAI ChatGPT was used to assist with manuscript editing,
including wording, condensation, and LaTeX organization.
ChatGPT was also used to assist with related-work and citation formatting. Cursor with Claude Opus 5.5 was used to generate
and refine code for implementation, experimental analysis,
and figure preparation; the resulting code and outputs
were reviewed and validated by the authors. GPT-5.3 was used to propose rule-based mappings for out-of-vocabulary anatomical phrases; these proposals were manually reviewed and verified by the authors. MedGemma 27B~\cite{sellergren2026medgemmatechnicalreport} was used to generate silver progression, anatomy, and sentence-level annotations from paired radiology reports. Qwen3-32B-AWQ~\cite{qwen3technicalreport} was used to generate independent annotations for cross-model consistency analysis. The authors designed the methodology, conducted the experiments, and reviewed and verified the scientific claims and reported results.

{
    \small
    \bibliographystyle{ieeenat_fullname}
    \bibliography{refs}
}


\clearpage
\onecolumn
\raggedbottom
\appendix
\section*{Appendix}
\renewcommand{\thetable}{A\arabic{table}}
\setcounter{table}{0}
\renewcommand{\thefigure}{A\arabic{figure}}
\setcounter{figure}{0}

\section{Dataset Characteristics}
\label{sec:supp_dataset_characteristics}

Table~\ref{tab:supp_dataset_characteristics} summarizes available demographic and examination-interval information from the earlier dataset characterization, alongside related CXR resources. Ages and percentages are descriptive cohort summaries; missing or differently defined metadata limit direct comparisons across resources. Scanner characteristics are not consistently available in the released datasets. In particular, CheXpert examination intervals are omitted because de-identified dates preserve study ordering but not reliable calendar-day differences.

\begin{table}[H]
\centering\small
\setlength{\tabcolsep}{5pt}
\caption{Available dataset characteristics. Age is mean $\pm$ standard deviation, and intervals are medians between consecutive examinations where date metadata permit calculation. The multi-source CheXTemporal cohorts aggregate CheXpert, MIMIC-CXR, and ReXGradient. Statistics are descriptive and should not be interpreted as matched cohorts.}
\label{tab:supp_dataset_characteristics}
\resizebox{\textwidth}{!}{%
\begin{tabular}{lcccl}
\toprule
Dataset & Age (years) & Male (\%) & Median interval & Institution / source \\
\midrule
CheXTemporal gold & $64.4\pm18.4$ & 52.8 & 13 days & Multi-source: CheXpert, MIMIC-CXR, ReXGradient \\
CheXTemporal silver & $63.5\pm17.6$ & 55.3 & 7 days & Multi-source: CheXpert, MIMIC-CXR, ReXGradient \\
CheXpert & $60.3\pm18.5$ & 55.4 & --- & Stanford \\
MIMIC-CXR & $56.9\pm19.5$ & 47.6 & 3 days & BIDMC \\
ReXGradient & $45.6\pm24.3$ & 48.4 & 11 days & Three U.S. health systems (79 sites) \\
MS-CXR-T & $65.4\pm16.6$ & 50.5 & 1 day & MIMIC-derived \\
Chest ImaGenome gold & $62.9\pm17.3$ & 52.3 & 3 days & MIMIC-derived \\
Chest ImaGenome silver & $56.8\pm19.5$ & 45.1 & 3 days & MIMIC-derived \\
\bottomrule
\end{tabular}}
\end{table}

\section{Gold and Silver Annotation Schemes}

Table~\ref{tab:annotation_schema} summarizes the complementary supervision provided by the gold and silver datasets. Gold annotations are based on radiologist assessment of paired images and include region-level progression labels and cross-time pathology correspondence. Silver annotations are derived from paired reports and provide finding-level progression labels, anatomical locations, and sentence-level evidence.

\begin{table}[H]
\centering
\caption{CheXTemporal annotation schemes. Gold supplies expert pathology localization and correspondence, whereas silver supplies report-derived anatomical grounding.}
\label{tab:annotation_schema}
\setlength{\tabcolsep}{8pt}
\begin{tabular}{@{}p{0.30\linewidth}p{0.29\linewidth}p{0.29\linewidth}@{}}
\toprule
 & Gold & Silver \\
\midrule
Label source & Radiologists; paired CXRs & MedGemma; paired reports \\
Progression unit & Region label, grouped by pair--finding & Pair--finding \\
Spatial labels & Pathology boxes with cross-time correspondence & Anatomy labels and CXAS masks \\
Report sentences & Not required for annotation & Static/dynamic evidence \\
\bottomrule
\end{tabular}
\end{table}

\section{Gold Annotation Counting Units}

The gold dataset supports multiple evaluation protocols that operate on different annotation units. Table~\ref{tab:gold_counting_units} distinguishes region-level progression labels, pair--finding groups, and the subsets used for progression-specific spatial grounding.

\begin{table}[H]
\centering
\caption{Gold counting units. The counts represent different annotation groupings and evaluation subsets, not interchangeable sample sizes.}
\label{tab:gold_counting_units}
\setlength{\tabcolsep}{5pt}
\begin{tabular}{@{}lr@{}}
\toprule
Counting unit & Count \\
\midrule
Progression labels & 1,787 \\
Pair--finding groups & 1,497 \\
Pair--finding--label annotations in grounding subset & 1,565 \\
Annotations with a current-study box & 1,489 \\
\bottomrule
\end{tabular}
\end{table}

\section{LLM Prompting for Silver Dataset Creation}
\begin{figure}[H]
\centering
\begin{minipage}{0.85\linewidth}
\scriptsize
\begin{Verbatim}[breaklines=true,breakanywhere=true]
You are an expert radiologist comparing chest X-ray reports.

Classify progression of ONE finding using EXACTLY one label:

New
Worse
Stable
Improved
Resolved

Definitions:

New
Finding present in current report but absent in prior.

Worse
Finding present in both reports but increased severity.

Stable
Finding present in both without meaningful change.
Words like "stable", "persistent", "unchanged", or
"again seen" imply Stable.

Improved
Finding present in both but decreased severity.

Resolved
Finding present previously but absent now.

IMPORTANT RULES

Base the progression decision only on statements about the specified finding.

If the finding involves multiple anatomical locations,
include all applicable anatomies separated by commas using
only terms from the anatomy list.

Choose anatomy EXACTLY from:

left lung, right lung,
cardiac silhouette, mediastinum,
left lower lung, right lower lung,
right hilar structures, left hilar structures,
upper mediastinum,
left costophrenic angle,
right costophrenic angle,
left mid lung, right mid lung,
aortic arch,
right upper lung, left upper lung,
right hemidiaphragm,
right clavicle, left clavicle,
left hemidiaphragm,
right apex, trachea, left apex,
carina, svc, right atrium,
cavoatrial junction,
abdomen, spine.

Output ONLY raw JSON.

{
  "finding":"...",
  "progression":"...",
  "anatomy":"...",
  "reasoning":"...",
  "evidence":[ "...", "..." ]
}
\end{Verbatim}
\end{minipage}
\caption{System prompt used for finding-level progression classification.}
\label{fig:findings_system_prompt}
\end{figure}

\begin{figure}[H]
\centering
\begin{minipage}{0.85\linewidth}
\scriptsize
\begin{Verbatim}[breaklines=true,breakanywhere=true]
You are an expert radiologist labeling sentences in the
IMPRESSION and FINDINGS sections of a chest X-ray report.

Label EVERY sentence describing an imaging finding.

Labels:

static
Sentence describes the current finding only.

dynamic
Sentence implies comparison or temporal change.

IMPORTANT RULES

Words such as stable, persistent, unchanged, changed,
again seen, increased, decreased, resolved, improved,
interval change, remaining, now, new, prior,
developing, re-demonstrated, still, or continued
MUST be labeled dynamic.

If a sentence contains an imaging finding,
it MUST appear in the output.

Ignore administrative text such as accession numbers
or anonymization statements.

Phrases like "no new ..." ARE dynamic.

Output ONLY raw JSON:

{
 "sentences":[
   {
     "study_id":"study1",
     "sentence":"Example sentence",
     "label":"static"
   }
 ]
}
\end{Verbatim}
\end{minipage}
\caption{System prompt used for sentence-level static/dynamic labeling.}
\label{fig:sentence_system_prompt}
\end{figure}
\subsection{MedGemma Prompts}
The silver dataset was constructed using the instruction-tuned \texttt{google/medgemma-27b-text-it} model with two categories of prompts, as shown in Figs.~\ref{fig:findings_system_prompt} and~\ref{fig:sentence_system_prompt}. Finding-level prompts asked the model to classify temporal progression for a specific radiographic finding using one of five labels: New, Worse, Stable, Improved, or Resolved. The anatomical labels were referenced from ChestImaGenome and further revised by a radiologist. Second, sentence-level prompts labeled report sentences as either static or dynamic depending on whether the sentence implied temporal comparison or progression. Outputs were parsed into structured JSON files. Experiments were run on four NVIDIA H100 80GB GPUs for distributed inference.
\subsection{Quality Filtering}
Although the structured prompt restricts MedGemma to a fixed vocabulary
of progressions and anatomical regions, free--form generation
occasionally yields malformed JSON or labels outside these label sets.
To ensure that downstream supervision uses a consistent label
space, we apply a deterministic post--hoc audit to every model output.
Each predicted record is first validated for JSON well--formedness;
records that fail to parse, or whose progression field falls outside
\{new, worse, stable, improved, resolved\}, are flagged as invalid and
discarded before dataset use. The finding label is supplied to the model as
part of the prompt (each record is conditioned on a target finding from
the silver--standard label set), so no auditing is required for that
field. For the remaining records, we check the anatomy field against
the closed vocabulary of 30 anatomical regions defined by the prompt.
Out--of--vocabulary values are normalized to the closest in-vocabulary
label using a small set of rule-based regular--expression mappers (e.g.,
``bibasilar'' to ``left lower lung, right lower lung''; ``perihilar''
to ``left hilar structures, right hilar structures''; ``retrocardiac''
to ``cardiac silhouette''). The mapping rules were proposed by GPT--5.3
from the set of observed out--of--vocabulary phrases and then verified
manually against the prompt's label inventory. Anatomical strings are
tokenized on commas, mapped part--wise, and de--duplicated; laterality
(``left''/``right'') is preserved when present and otherwise expanded
to both sides. Less than 1\% of outputs were thrown out entirely, and
around 7\% had at least one anatomical phrase remapped
(Figure~\ref{fig:audit-mapping}).

\begin{figure}[H]
\centering
\small
\setlength{\tabcolsep}{4pt}
\renewcommand{\arraystretch}{1.05}
\begin{adjustbox}{max width=\linewidth}
\begin{tabular}{@{}p{0.46\linewidth} c p{0.42\linewidth}@{}}
\toprule
Predicted phrase contains\ldots & $\rightarrow$ & Mapped anatomical region \\
\midrule
apex, apical                          & $\rightarrow$ & \{left, right\} apex \\
upper, suprahilar                     & $\rightarrow$ & \{left, right\} upper lung \\
mid                                   & $\rightarrow$ & \{left, right\} mid lung \\
lower, base, basilar, bibasilar, bases & $\rightarrow$ & \{left, right\} lower lung \\
hilar, hilum, perihilar               & $\rightarrow$ & \{left, right\} hilar structures \\
mediast*, paratracheal, subcarinal    & $\rightarrow$ & mediastinum \\
cardiac, heart                        & $\rightarrow$ & cardiac silhouette \\
aorta                                 & $\rightarrow$ & aortic arch \\
svc, cavoatrial                       & $\rightarrow$ & svc \\
trachea                               & $\rightarrow$ & trachea \\
pleura*, costophrenic                 & $\rightarrow$ & \{left, right\} costophrenic angle \\
diaphragm                             & $\rightarrow$ & \{left, right\} hemidiaphragm \\
spine, vertebra                       & $\rightarrow$ & spine \\
clavicle                              & $\rightarrow$ & \{left, right\} clavicle \\
abdomen, bowel, stomach, liver        & $\rightarrow$ & abdomen \\
(default / unmatched)                 & $\rightarrow$ & left lung, right lung \\
\bottomrule
\end{tabular}
\end{adjustbox}
\caption{Rule--based mapping from out--of--vocabulary anatomical phrases to
the 30--region closed label space. Each generated record is first
validated for JSON well-formedness and a valid progression in \{new,
worse, stable, improved, resolved\}; invalid records are discarded.
Anatomy values that fall outside the prompt vocabulary are normalized
using the regular--expression rules above. The mapping rules were
drafted by GPT-5.3 from the set of observed out--of--vocabulary phrases
and then manually verified by the authors. Laterality cues
(``left''/``right'') in the generation are preserved when present;
otherwise the rule expands to both sides. Anatomy strings are split on
commas and mapped part-wise before de-duplication.}
\label{fig:audit-mapping}
\end{figure}
\subsection{Mask Filtering and Quality Control}
Mask quality control was automatic rather than per-case expert review. Rules for anatomical coverage, required classes, and implausible masks were developed with expert input, with limited overlay spot-checks during pipeline development. Individual masks were not manually accepted or rejected by radiologists.
\begin{figure}[H]
\centering
\setlength{\tabcolsep}{2pt}
\begin{adjustbox}{max width=\linewidth}
\begin{tabular}{cc}
\includegraphics[width=0.4\linewidth]{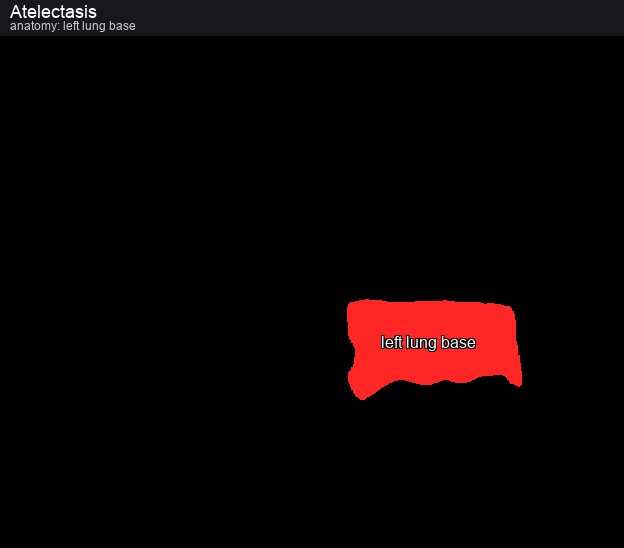} &
\includegraphics[width=0.4\linewidth]{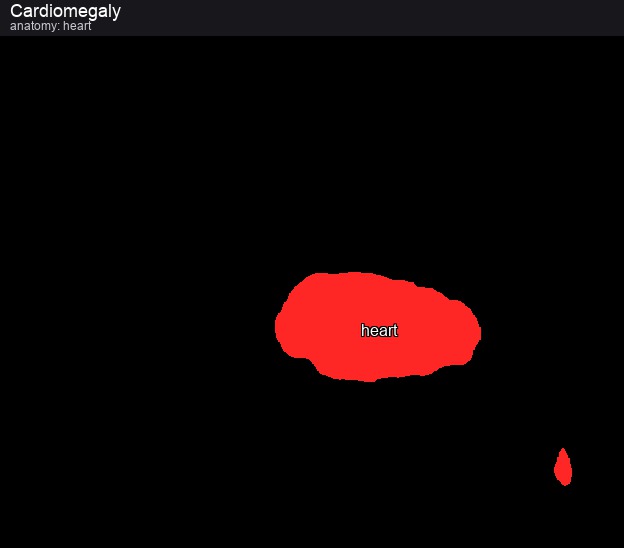} \\
\includegraphics[width=0.4\linewidth]{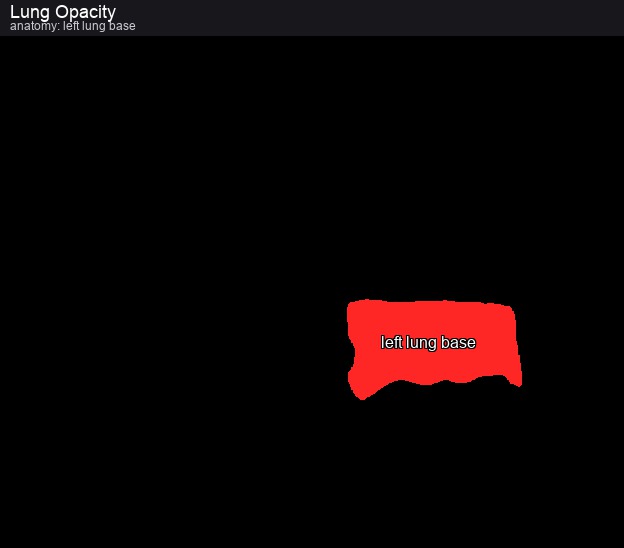} &
\includegraphics[width=0.4\linewidth]{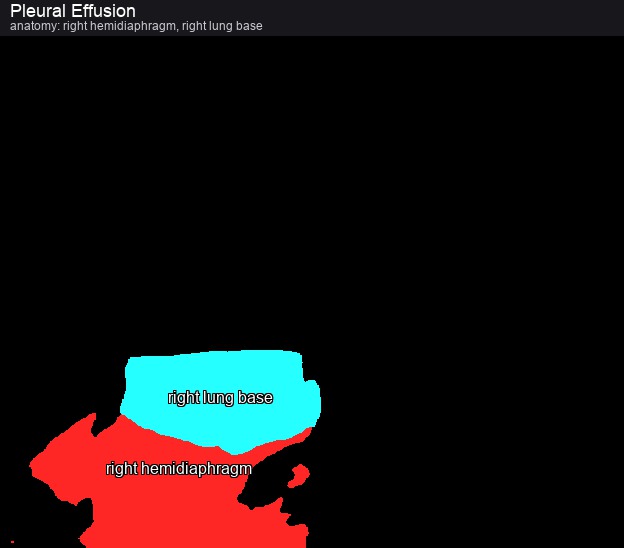}
\end{tabular}
\end{adjustbox}
\caption{Examples of anatomy--derived segmentation masks supplied as silver spatial supervision. Each panel shows the target finding and the anatomical region or regions retained after mapping report--derived anatomy mentions to CXAS segmentation classes.}
\label{fig:silver_segmentation_masks}
\end{figure}

Each MedGemma anatomy phrase is mapped to its corresponding
CXAS class, and phrases without a confident mapping are dropped rather than aliased to a coarser region. A study--level quality gate then discards any CXAS mask in which fewer than $8$ of $10$ major anatomy classes (lungs, heart, spine, clavicles, trachea, hemidiaphragms, aortic arch) carry annotations -- a single threshold that captures the dominant CXAS failure mode (cropped views, abdomen-included frames, extreme rotations) and is what gates ReXGradient out of the released mask set entirely. Surviving study masks are filtered down to the requested anatomy: we keep only annotations whose CXAS class is in the finding's mapped target set and drop the example outright if any required class is missing. The released JSON for each (study, finding) is therefore exactly the anatomical region MedGemma identified, which lets the masks be used directly as supervision for phrase-grounded localization. Figure~\ref{fig:silver_segmentation_masks} shows representative retained anatomy-derived masks; they indicate anatomical regions, not exact pathology boundaries.

\subsection{Manual Validation Protocol and Uncertainty}
\label{sec:supp_silver_validation}
\begin{figure}[H]
    \centering
    \includegraphics[width=0.69\linewidth]{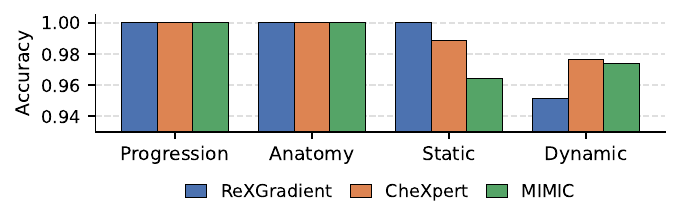}
    \caption{Manual analysis of $55$ randomly sampled reports per dataset in the silver dataset. Progression and anatomy accuracies are computed at the finding level; static and dynamic accuracies are computed at the sentence level.}
    \label{fig:manual}
\end{figure}
The manual silver validation sampled 55 longitudinal study pairs (paired reports) from each of CheXpert, MIMIC-CXR, and ReXGradient. The first three authors divided the sampled cases, and each case received one independent review. Reviewers assessed whether progression labels, anatomical labels, and static/dynamic sentence classifications were faithful to the paired report text, not whether the reports or labels were clinically correct on imaging. Two radiologists involved in gold annotation advised on prompt design and anatomical definitions and conducted limited checks during development; they did not independently validate the full sampled cohort. The segmentation masks were quality-filtered automatically, not reviewed individually by experts.

Table~\ref{tab:supp_silver_manual_ci} reproduces the 95\% Wilson intervals for the observed report-fidelity accuracies. Even where observed accuracy was 100\%, these intervals reflect uncertainty from finite sampling. This limited, single-reviewer report-based validation does not establish the clinical correctness of all automatically derived silver annotations; the cross-model agreement analysis in the main-text Silver Dataset subsection provides complementary consistency evidence rather than an expert reference standard.

\begin{table}[H]
\centering\small
\setlength{\tabcolsep}{7pt}
\caption{Manual silver annotation report-fidelity accuracy and 95\% Wilson confidence intervals from sampled paired reports. Progression and anatomy refer to finding-level labels; Static and Dynamic refer to sentence-level labels. These results are not expert image-level validation.}
\label{tab:supp_silver_manual_ci}
\begin{adjustbox}{max width=\linewidth}
\begin{tabular}{lccc}
\toprule
Dataset & Progression / anatomy & Static & Dynamic \\
\midrule
ReXGradient & 1.000 [0.958, 1.000] & 1.000 [0.987, 1.000] & 0.952 [0.907, 0.975] \\
CheXpert & 1.000 [0.935, 1.000] & 0.989 [0.938, 0.998] & 0.976 [0.946, 0.990] \\
MIMIC-CXR & 1.000 [0.935, 1.000] & 0.964 [0.929, 0.983] & 0.974 [0.934, 0.990] \\
\bottomrule
\end{tabular}
\end{adjustbox}
\end{table}

\subsection{Detailed Silver Cross-Model Agreement}
\label{sec:supp_silver_crossmodel}

Across 281,836 matched findings, progression-label agreement was 80.1\% (Cohen's $\kappa = 0.71$, substantial agreement), with class-specific agreement of 89.7\% for New, 86.7\% for Improved, 78.5\% for Worse, 77.2\% for Stable, and 76.2\% for Resolved. Agreement was consistent across CheXpert (80.2\%), MIMIC-CXR (79.9\%), and ReXGradient (81.8\%).

Most disagreements were not contradictory: 73.8\% were between Stable and a change label, typically reflecting differing thresholds for what constitutes interval change, and 17.6\% were between labels of the same direction (e.g., New vs.\ Worse). Opposite-direction disagreements (improvement vs.\ worsening) accounted for only 1.7\% of all findings. The pattern of cross-model disagreement on silver resembles inter-radiologist disagreement on gold: most disagreements concern whether a finding is stable or has changed (73.8\% on silver and 78\% on gold), rather than the direction of change.

Static/dynamic sentence classification achieved 90.3\% agreement ($\kappa = 0.80$) across 630,540 sentences, with 94.8\% for Static and 87.2\% for Dynamic. For anatomical annotations, fine-grained region labels achieved 57.4\% exact-set agreement, 64.2\% token F1, and 69.6\% any-overlap agreement. Agreement increased at the organ-side level (78.3\% exact match, 87.9\% token F1) and laterality level (84.9\% exact match, 93.3\% token F1, 96.7\% any-overlap).

\subsection{Detailed Gold Inter-Radiologist Agreement}
\label{sec:supp_gold_agreement}

Across 415 matched cases, unweighted five-class Cohen's $\kappa$ was 0.339 (95\% CI: 0.261--0.424) on 305 cases with a single label from each rater, with 54.8\% exact agreement; full label sets matched in 41.9\% of cases and overlapped in 62.4\%. Disagreements were concentrated between neighboring categories: linear- and quadratic-weighted $\kappa$ rose to 0.427 (0.342--0.509) and 0.504 (0.390--0.616), 78\% of disagreements were between \emph{stable} and a change label, 14\% were between classes with the same direction (\emph{new} vs.\ \emph{worse}, \emph{resolved} vs.\ \emph{improved}), and only 11 of 305 cases (3.6\%) received opposite directions (improvement vs.\ deterioration). Radiologists thus rarely disagree on the direction of change but often on whether a subtle change is present, which is reflected in higher agreement for worsening (Worse/New vs.\ other categories; $\kappa=0.544$) than for any change versus Stable ($\kappa=0.326$), and in class-specific $\kappa$ being highest for Worse (0.45) and lowest for Stable (0.26).

Across 497 matched spatial cases, mean best-match IoU was 0.431 (95\% CI: 0.412--0.450) for prior and 0.427 (0.409--0.444) for current studies; about 70\% of matched boxes reached IoU $\geq 0.3$ but only 39\% reached $\geq 0.5$, indicating that raters consistently mark the same region while differing in its exact extent, as expected for approximate boxes around often ill-defined opacities. Spatial agreement was highest for the well-delineated cardiac silhouette (Cardiomegaly, IoU $\approx 0.70$) and lowest for diffuse or subtle findings (Consolidation in prior and Pneumothorax in current studies, both $\approx 0.31$).

\section{Annotation Platform Instructions}
\begin{figure}[H]
    \centering
    \includegraphics[width=1\linewidth]{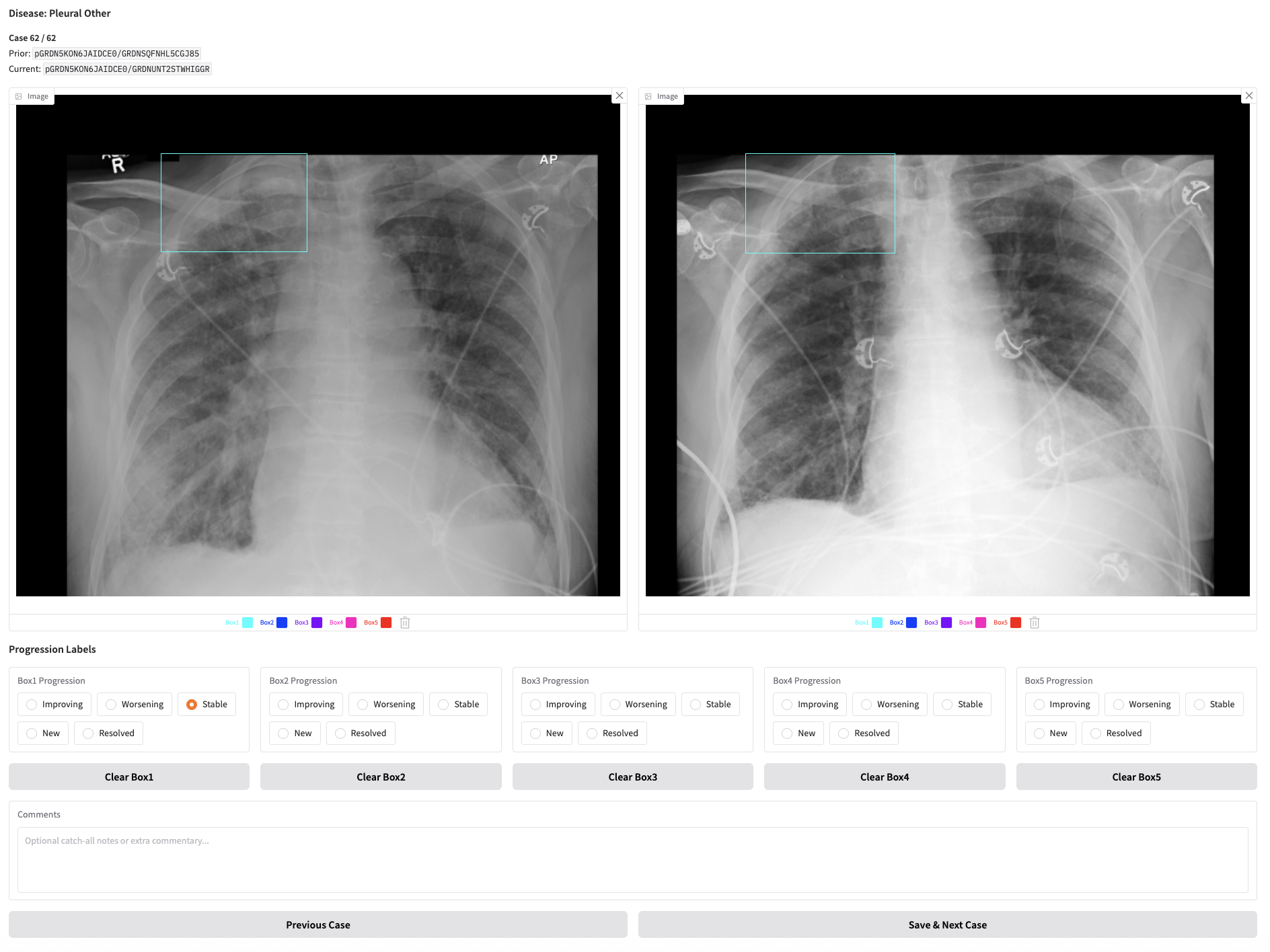}
    \caption{Example annotation window on the Gradio platform.}
    \label{fig:gradio_platform}
\end{figure}
Figure~\ref{fig:gradio_platform} shows the Gradio interface used to collect gold annotations. The following instructions were provided to radiologist annotators.

\paragraph{Workflow rationale and cross-time correspondence.}
The interface was developed iteratively with participating radiologists, who preferred one target finding at a time while keeping both full radiographs visible. The upper limit of five regions was selected as a practical balance between capturing clinically relevant multifocal findings and maintaining a manageable annotation workload and reliable region correspondence. For extensive or diffuse localized abnormalities, annotators could draw coarse boxes around clinically relevant regions rather than marking every small focus. The same numbered box identity denotes a location in both examinations when the finding persists; new or resolved locations may appear on only one image. These correspondences were assigned by visual radiologist assessment, without image registration or automated alignment. The instructions below are preserved from the annotation interface.

\subsection{Dataset Goal}
Tracking how clinical findings change over time on chest X-rays is central to radiology practice. Radiologists routinely compare a current image against a prior study to assess whether a finding has improved, worsened, remained stable, newly appeared, or resolved. This temporal reasoning is critical for clinical decision--making, yet existing datasets for training and evaluating AI models in this area are limited. They typically cover only a small set of findings, provide only image-level progression labels without spatial localization, and use a narrow set of progression categories (Improved, Stable, Worse).

This dataset aims to address these gaps. We are building a temporally annotated chest X-ray dataset that pairs prior and current frontal chest X-rays and provides, for each clinical finding:

\begin{itemize}
\item \textbf{Spatial localization:} Bounding boxes marking where the finding appears in each image.
\item \textbf{Cross-image correspondence:} Explicit matching of each annotated location across the prior and current images.
\item \textbf{Progression labels:} A five-class label per location: Improved, Worse, Stable, New, or Resolved.
\end{itemize}

The dataset covers 11 clinical findings:
\begin{itemize}
\item Enlarged Cardiomediastinum
\item Cardiomegaly
\item Lung Opacity
\item Lung Lesion
\item Edema
\item Consolidation
\item Pneumonia
\item Atelectasis
\item Pneumothorax
\item Pleural Effusion
\item Pleural Other
\end{itemize}

Edema is included with progression labels only (no bounding boxes) due to its diffuse nature. Your annotations will directly support the development and evaluation of vision-language models that reason about temporal change in medical imaging.

\subsection{Platform Overview}
The annotation platform presents one case at a time. Each case consists of a single clinical finding and a pair of chest X-ray images.

\begin{itemize}
\item \textbf{Image display area.} Two frontal chest X-ray images are shown side by side. The left image is the Prior (earlier study) and the right image is the Current (later study).

\item \textbf{Disease label.} The finding to be annotated is displayed at the top of the interface (e.g., "Disease: Pneumonia"). A single pair of images may contain multiple findings, but the platform presents one finding at a time for annotation. This means you may see the same image pair more than once, each time with a different finding to evaluate. Focus only on the finding shown at the top for the current annotation iteration.

\item \textbf{Bounding box tools (Box 1--Box 5).} Up to five bounding boxes are available for drawing. Each box represents one distinct location where the finding appears in the image. You can draw boxes on either or both images. The same box label (e.g., Box 1) should refer to the same disease location across both images — for example, if Box 1 marks a region of pneumonia in the right lower lung on the prior image, Box 1 on the current image should mark that same pneumonia in the right lower lung.

\item \textbf{Progression label dropdown.} Each box has an associated dropdown where you select a progression label: Improved, Worse, Stable, New, or Resolved. Every box that has been drawn must have a label assigned before saving.

\item \textbf{Comments field.} A free--text field for optional notes — for example, diagnostic uncertainty, image quality issues, or additional observations.

\item \textbf{Navigation buttons.}
    \begin{itemize}
    \item Previous Case: Returns to the last saved case for review or correction.
    \item Save \& Next Case: Saves all annotations (boxes, labels, comments) and loads the next case. The system will prompt you if any drawn box is missing a progression label.
    \end{itemize}
\end{itemize}

\subsection{Annotation Steps}

\paragraph{Step 1: Review the Images and Finding Label}
Begin by reading the disease label at the top of the interface. Then examine both the prior and current images, focusing specifically on that finding. Assess whether and where the finding is present in each image before drawing any boxes.

\textbf{Example:}\\
Disease: Pneumonia\\
Important: although only one disease is given, it may appear in multiple parts of the chest X-ray.

\paragraph{Step 2: Draw Bounding Boxes}
You may draw up to 5 boxes (Box1 - Box5). Each box represents one distinct location where the disease appears. If the appearance of a disease in a particular location is very fine-grained, you are allowed to draw a general bounding box around that location of the disease.

In general, it is preferable to limit the number of boxes drawn in each image, so only increase the number of boxes if it is necessary for capturing the full extent of the disease.

\textbf{Example:}
\begin{itemize}
\item Right lower lung → Box1
\item Left upper lung → Box2
\end{itemize}

Each box tracks that specific region over time.

\paragraph{Step 3: Make Boxes Correspond Across Images}
Each box label should refer to the same disease location in both images.

\textbf{Example:}
\begin{itemize}
\item Box1 → pneumonia in right lower lung (prior)
\item Box1 → pneumonia in right lower lung (current)
\item Box2 → pneumonia in left upper lung (prior)
\item Box2 → pneumonia in left upper lung (current)
\end{itemize}

Box labels represent disease locations, not different diseases.

\paragraph{Step 4: When the Disease Appears or Disappears}
Sometimes a disease location may appear only in one image.

\begin{itemize}
\item New disease location\\
Draw a box only on the current image and label the progression New.
\item Resolved disease location\\
Draw a box only on the prior image and label the progression Resolved.
\end{itemize}

\paragraph{Step 5: Edema Exception}
Edema does NOT require a bounding box.\\
If edema is present in the case, you do not need to draw boxes for it.\\
Boxes are only used for localized disease regions.

\paragraph{Step 6: Assign Progression Labels}
For each box used, select a progression label.

Available options:
\begin{itemize}
\item Improved: disease decreased
\item Worse: disease increased
\item Stable: no significant change
\item New: appears only in the current image
\item Resolved: present before but gone now
\end{itemize}

Every box that you draw must have a progression label.

\paragraph{Step 7: Comments (Optional)}
You may add notes in the Comments box.\\
Examples include uncertainty in interpretation or additional observations.\\
This field is optional.

\paragraph{Step 8: Save the Case}
Click Save \& Next Case.

The platform will:
\begin{itemize}
\item Save the bounding boxes
\item Save the progression labels
\item Load the next case
\end{itemize}

If a progression label is missing for a used box, the system will prompt you to complete it before continuing.

\paragraph{Step 9: Navigating Cases}
\begin{itemize}
\item Previous Case: Go back to the previous case.
\item Save \& Next Case: Save the current annotations and move to the next case.
\end{itemize}

\subsection{Key Rules Summary}
\begin{itemize}
\item Each page prompts for the annotation of one disease at a time, even if the prior and current chest X-ray pairs contain multiple diseases.
\item The disease may appear in multiple locations in the lungs.
\item Each box represents one disease location.
\item The same box label must refer to the same location in both images.
\item Every used box must have a progression label.
\item Edema does not require bounding boxes.
\end{itemize}
\section{Progression Label Synonym Mapping}
The progression classification pipeline maps each candidate
sentence to one of five classes (stable, improved, worse, new,
resolved) by scoring it against a hand-curated bank of $37$
synonymous direction-of-change phrases:
stable: \{constant, stable, unchanged\};
improved: \{better, decreased, decreasing, improved, improving,
reduced, smaller\};
worse: \{bigger, developing, enlarged, enlarging, greater,
growing, increased, increasing, larger, progressing, progressive,
worse, worsened, worsening\};
new: \{new, newly developed, newly appeared, newly seen, appeared,
emerged\};
resolved: \{resolved, resolving, cleared, disappeared, no longer
present, no longer seen, completely resolved\}.
Each phrase is embedded with the template \texttt{"\{disease\} is \{phrase\}"} and the sentence is assigned the class whose bank yields the highest cosine similarity.

Temporal models encode image pairs jointly, while single-image models use the normalized difference between current- and prior-image embeddings.

\section{Prog-JEPA Training and Implementation Details}
\label{sec:supp_training_implementation}

\noindent\textbf{Training and implementation.}
Prog-JEPA trains its predictor with frozen image and text encoders, while fine-tuned Prog-JEPA additionally updates the image encoder. Both variants use the same silver training data. The progression-aware text adapter is trained separately, and class-score offsets are selected on silver validation.

In the frozen-encoder configuration, approximately 0.75B predictor parameters are trained with learning rate $2\times10^{-5}$. Fine-tuned Prog-JEPA initializes from Prog-JEPA and additionally trains the image encoder at $2\times10^{-6}$, keeping its BatchNorm statistics and dropout fixed. Both variants use AdamW (weight decay 0.01), 3\% linear warmup followed by linear decay, gradient clipping at 1.0, batch size 32 across four GPUs, and gradient checkpointing. Images are resized to 512 pixels and center-cropped to $448\times448$; identical augmentations are applied to each image pair (rotation $\pm30^\circ$, shear $\pm15^\circ$, brightness and contrast in $[0.8,1.2]$).

The progression-aware text adapter trains only its LoRA parameters on 220 templated sentences at learning rate $10^{-3}$. At inference, all class offsets are zero except Resolved, which comprises 1.1\% of training data; its validation-selected offset is 0.125 for Prog-JEPA and 0.1075 for fine-tuned Prog-JEPA. Resolved is excluded from the three-class benchmarks, so these offsets do not affect external three-class scores.

\noindent\textbf{Alternative-training protocols.}
Dynamic Prog-JEPA uses dynamically labeled report sentences as training targets, while Laterality Prog-JEPA uses agreement on anatomy between two silver annotation models. Dynamic uses the original frozen text encoder, whereas Laterality uses the frozen text-LoRA encoder. Their Resolved-score offsets are selected on silver validation.

Specifically, Dynamic Prog-JEPA samples one dynamically labeled report sentence per image pair anew each epoch and learns with cross-GPU in-batch InfoNCE. At evaluation, a shared pair embedding is matched to the 37-phrase progression bank. Laterality Prog-JEPA uses examples with agreement on anatomy between the two silver annotation models, representing 73.0\% of training examples. Both variants first train with the image encoder frozen and then unfreeze it at one-tenth the predictor learning rate.

Unlike the original five-way implementation, both alternative variant trainers synchronize gradients across four GPUs (global batch size 32), so the optimization protocols differ in effective batch size. These differences, together with the distinct text encoders and target formulations, mean that comparisons among supervision variants are not strictly controlled ablations of the target representation alone.

\section{Complete Progression Classification Results}
\label{sec:supp_progression_tables}

The tables below reproduce the complete progression classification comparisons, including frozen-image-encoder and unshifted Laterality Prog-JEPA configurations omitted from the main-paper tables. The main paper consistently displays fine-tuned configurations; all results use the same evaluation protocols.

\begin{table}[H]
\centering
\caption{Complete progression classification on CheXTemporal gold (1,497 groups). Combined is mean label-set Jaccard; F1 scores use 1,239 single-label groups.}
\label{tab:vljepa_gold_full}
\setlength{\tabcolsep}{6pt}\resizebox{0.75\textwidth}{!}{%
\begin{tabular}{lccccccc}\toprule
Model & Combined & Macro F1 & Impr. & Stable & Wors. & New & Resolved\\\midrule
BioViL~\cite{Boecking_2022} & .243 & .182 & .137 & .062 & .430 & .146 & .133\\
BiomedCLIP~\cite{zhang2025biomedclipmultimodalbiomedicalfoundation} & .223 & .150 & .115 & .104 & .354 & .060 & .117\\
BioViL-T~\cite{bannur2023learningexploittemporalstructure} & .307 & .249 & .297 & .145 & .417 & .212 & .175\\
TILA~\cite{ko2026temporalinversionlearninginterval} & .261 & .208 & .230 & .107 & .373 & .202 & .129\\
ALTA~\cite{Lian_2025} & .256 & .213 & .318 & .132 & .278 & .176 & .159\\\midrule
Supervised & .457 & .393 & .463 & .515 & .488 & .272 & .228\\
Prog-JEPA (5-way) & .471 & .393 & .433 & .583 & .406 & .305 & .239\\
Prog-JEPA (5-way, fine-tuned) & \textbf{.480} & .394 & .370 & \textbf{.612} & .436 & \textbf{.308} & .244\\
Dynamic Prog-JEPA & .325 & .270 & .395 & .202 & .404 & .193 & .155\\
Dynamic Prog-JEPA (fine-tuned) & .331 & .276 & .394 & .205 & .419 & .200 & .164\\
Laterality Prog-JEPA & .452 & .372 & .405 & .572 & .370 & .280 & .233\\
Laterality Prog-JEPA (fine-tuned) & .459 & .385 & .403 & .577 & .414 & .287 & .243\\\midrule
Laterality Prog-JEPA (unshifted) & .463 & .382 & .421 & .572 & .374 & .300 & .245\\
Laterality Prog-JEPA (fine-tuned, unshifted) & .477 & \textbf{.397} & .413 & .585 & .418 & .305 & \textbf{.263}\\\bottomrule
\end{tabular}}
\end{table}

\begin{table}[H]
\centering
\caption{Complete progression classification on MS-CXR-T (339 pairs). All entries are F1 except accuracy; all models use the same patient-filtered evaluation pairs.}
\label{tab:vljepa_mscxrt_full}
\scriptsize
\setlength{\tabcolsep}{3pt}
\begin{tabular}{@{}lccccc@{}}
\toprule
Model & Macro F1 & Impr. & Stable & Wors. & Acc.\\
\midrule
BioViL~\cite{Boecking_2022} & .381 & .374 & .137 & .632 & .460\\
BiomedCLIP~\cite{zhang2025biomedclipmultimodalbiomedicalfoundation} & .365 & .279 & .184 & .631 & .460\\
BioViL-T~\cite{bannur2023learningexploittemporalstructure} & .372 & .278 & .177 & .660 & .487\\
TILA~\cite{ko2026temporalinversionlearninginterval} & .420 & .359 & .248 & .654 & .490\\
ALTA~\cite{Lian_2025} & .331 & .321 & .161 & .510 & .366\\
Supervised & .420 & .433 & .120 & .707 & .513\\
Prog-JEPA (5-way) & .492 & .387 & .443 & .645 & .510\\
Prog-JEPA (5-way, fine-tuned) & \textbf{.512} & .402 & .410 & \textbf{.724} & \textbf{.549}\\
Dynamic Prog-JEPA & .446 & .400 & .233 & .705 & .522\\
Dynamic Prog-JEPA (fine-tuned) & .471 & .432 & .278 & .702 & .534\\
Laterality Prog-JEPA & .502 & .419 & .474 & .614 & .513\\
Laterality Prog-JEPA (fine-tuned) & .481 & .422 & .350 & .671 & .507\\
\bottomrule
\end{tabular}
\end{table}

\begin{table}[H]
\centering
\caption{Complete progression classification on Chest ImaGenome (348 pairs). All entries are F1 except accuracy; all models use the same patient-filtered evaluation pairs.}
\label{tab:vljepa_imagenome_full}
\scriptsize
\setlength{\tabcolsep}{3pt}
\begin{tabular}{@{}lccccc@{}}
\toprule
Model & Macro F1 & Impr. & Stable & Wors. & Acc.\\
\midrule
BioViL~\cite{Boecking_2022} & .322 & .297 & .155 & .513 & .351\\
BiomedCLIP~\cite{zhang2025biomedclipmultimodalbiomedicalfoundation} & .276 & .140 & .179 & .511 & .351\\
BioViL-T~\cite{bannur2023learningexploittemporalstructure} & .372 & .362 & .225 & .528 & .405\\
TILA~\cite{ko2026temporalinversionlearninginterval} & .336 & .293 & .257 & .458 & .353\\
ALTA~\cite{Lian_2025} & .266 & .276 & .011 & .510 & .273\\
Supervised & .509 & .388 & .521 & .619 & .526\\
Prog-JEPA (5-way) & .552 & .420 & .647 & .587 & .583\\
Prog-JEPA (5-way, fine-tuned) & \textbf{.586} & .403 & .695 & \textbf{.659} & \textbf{.632}\\
Dynamic Prog-JEPA & .353 & .393 & .093 & .572 & .397\\
Dynamic Prog-JEPA (fine-tuned) & .333 & .384 & .063 & .552 & .382\\
Laterality Prog-JEPA & .543 & .428 & .686 & .515 & .583\\
Laterality Prog-JEPA (fine-tuned) & .547 & .387 & \textbf{.697} & .557 & .598\\
\bottomrule
\end{tabular}
\end{table}

\subsection{Detailed Ablation and Alternative Supervision Results}
\label{sec:supp_ablation_results}

\paragraph{Gold classification.}
Dynamic Prog-JEPA reaches .325 combined (.331 after image fine-tuning), above the zero-shot baselines but well below the five-way model. Shifted laterality Prog-JEPA obtains .452 combined (.459 fine-tuned), versus .471 (.480 fine-tuned) for five-way Prog-JEPA. In 2,000 paired bootstrap resamples, the shifted fine-tuned laterality variant trails the fine-tuned five-way model by .021 combined (95\% CI: -.038 to -.003).

Without the silver-selected Resolved offset, laterality improves to .463 (.477 fine-tuned) combined; these unshifted results are secondary, and the fine-tuned unshifted variant improves Improved F1 by .044 (95\% CI: .003--.085) relative to fine-tuned five-way Prog-JEPA. The silver-selected shift can reduce gold performance for laterality because it increases Resolved predictions. Worse remains a weakness for both laterality stages. Side-readout accuracy on silver validation is .736 and .759 for frozen and fine-tuned laterality, respectively, compared with .546 for the majority bilateral class; gold has no side labels.

\paragraph{External transfer.}
Dynamic Prog-JEPA fine-tuned transfers to MS-CXR-T (macro F1 .471), exceeding all zero-shot baselines except TILA; it is statistically tied with TILA, the supervised baseline, and both five-way models, but exceeds BioViL-T by .099 (95\% CI: .038--.158). On Chest ImaGenome, however, dynamic fine-tuning reaches only .333 macro F1, below the supervised and both five-way models; it predicts Stable for just 6 of 348 pairs (Stable F1 .063). Frozen and fine-tuned dynamic variants are statistically tied on both benchmarks.

Laterality Prog-JEPA is statistically tied with the corresponding five-way model in macro F1 on both benchmarks: frozen differences are +.011 on MS-CXR-T and -.009 on Chest ImaGenome, and fine-tuned differences are -.031 (95\% CI: -.075 to +.013) and -.039 (95\% CI: -.090 to +.006), respectively. Worse is the consistent trade-off: fine-tuned laterality loses .102 F1 on Chest ImaGenome (95\% CI: -.171 to -.035) and .053 on MS-CXR-T (95\% CI: -.106 to +.001). The Resolved offset does not affect either external benchmark because evaluation is restricted to Improved, Stable, and Worse.
\section{Prog-JEPA Architecture Details}
\label{sec:supp_prog_jepa_architecture}

This section provides the implementation details of the temporal image encoder, target text encoder, and predictor used in Prog-JEPA. The main architectural components and learning objective are described in Sec.~\ref{sec:prog_jepa}.

\begin{figure}[H]
    \centering
    \includegraphics[width=0.85\linewidth]{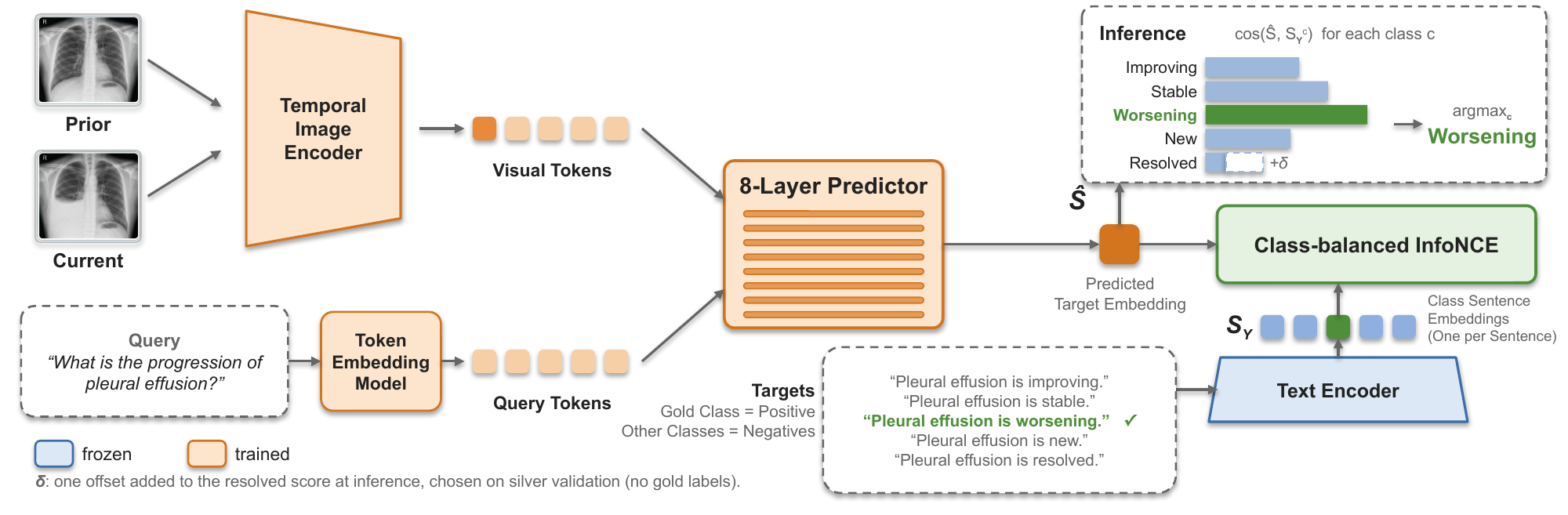}
    \caption{Overview of Prog-JEPA. Paired prior and current chest radiographs are encoded using BioViL-T, and a finding-conditioned transformer predictor maps temporal image features and a finding query into a progression-aware text embedding space. Training aligns the predicted embedding with the target progression sentence embedding. At inference, progression is determined by similarity to candidate class embeddings.}
    \label{fig:model_architecture}
\end{figure}

\paragraph{Temporal image encoder.}
We use the BioViL-T~\cite{bannur2023learningexploittemporalstructure} temporal image encoder, comprising a ResNet-50 backbone and a transformer fusing current-image features with prior-image-conditioned differences. It returns an $\ell_2$-normalized global vector $g\in\mathbb{R}^{128}$ and patch grid $P\in\mathbb{R}^{196\times128}$ at $14\times14$ resolution. Each patch feature combines current-image content and prior-conditioned difference information.

\paragraph{Target text encoder.}
We use BioViL-T's CXR-BERT text encoder and its 128-dimensional projection head. A rank-8 LoRA adapter ($\alpha=16$) is applied to query and value matrices in each self-attention layer. For finding $f$ and class $k$, the target sentence ``$f$ is $k$.'' is encoded as a normalized vector $s_k\in\mathbb{R}^{128}$. Both the base text encoder and the adapted weights are frozen during Prog-JEPA training.

\paragraph{Predictor.}
The predictor is initialized from the last eight of sixteen transformer layers of Llama-3.2-1B~\cite{meta2024llama32modelcard}, with hidden dimension 2048 and causal masking removed. The 197 visual tokens $[g;P]$ are projected to 2048 dimensions. We concatenate them with up to 64 pretrained Llama token embeddings for the query ``What is the progression of $f$?''; learned visual/query type embeddings, rotary positions, and padding masks are applied. Final hidden states pass through the Llama RMSNorm and an additional LayerNorm, are averaged over valid tokens, projected to 128 dimensions, and $\ell_2$-normalized:
\begin{equation}
\hat{s}=\frac{W_o\bigl(\frac{1}{N}\sum_i\operatorname{LN}(h_i)\bigr)}{\left\|W_o\bigl(\frac{1}{N}\sum_i\operatorname{LN}(h_i)\bigr)\right\|_2}.
\end{equation}
The predictor jointly attends to all patches and the finding query, but produces a single pair-level representation.

\section{Visualization of Learned Text Embeddings}
\label{app:text_embeddings}

\begin{figure}[H]
    \centering
    \includegraphics[width=\textwidth]{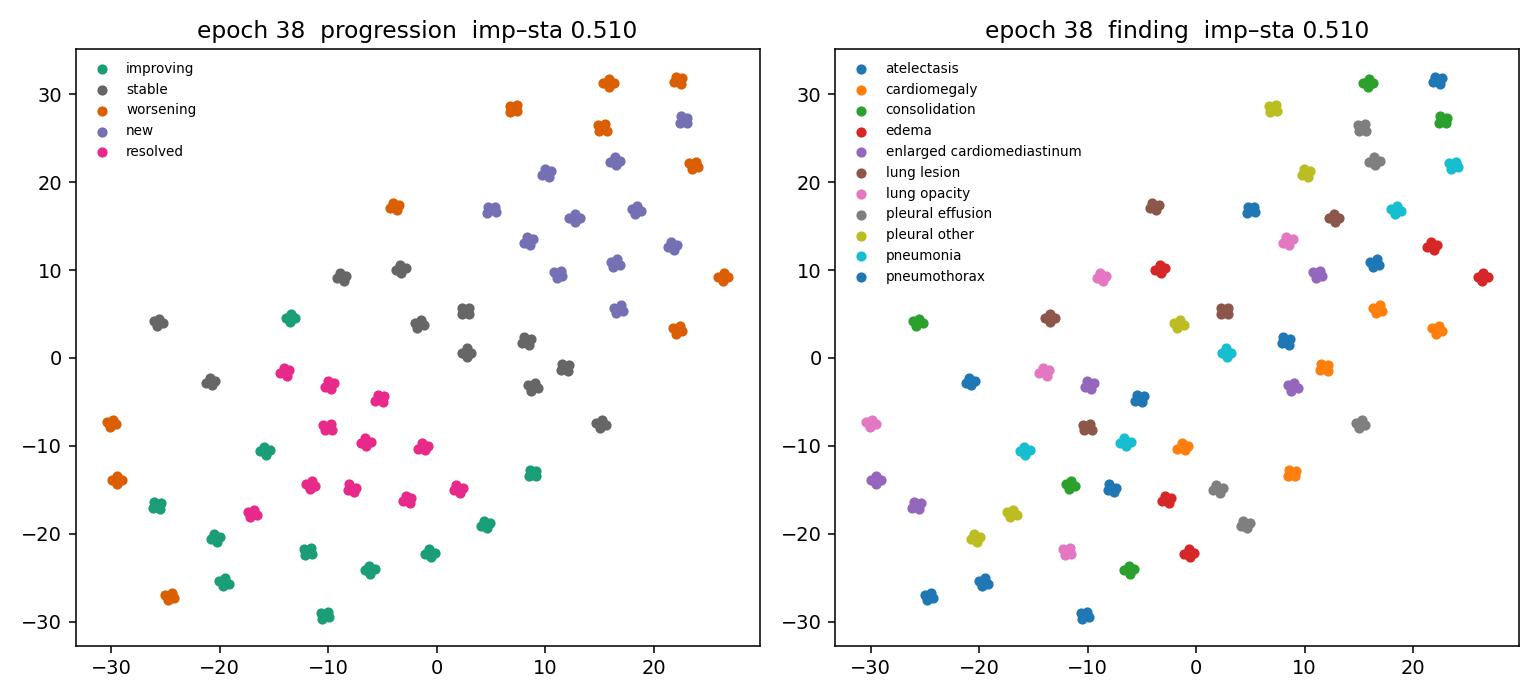}
    \caption{\textbf{Progression-aware text embedding visualization.} Embeddings after fine-tuning, colored by progression category (left) and radiographic finding (right). Progression categories exhibit more distinct spatial organization than finding identities.}
    \label{fig:text_embeddings}
\end{figure}

Figure~\ref{fig:text_embeddings} visualizes the learned text embeddings after progression-aware fine-tuning of the original BioViL-T text encoder. Compared to the original encoder, which primarily captures radiographic finding semantics, our progression-aware encoder encourages separation based on temporal change. The resulting embeddings exhibit clearer organization by progression category, with improved and resolved cases concentrated in the lower region and new and worse cases predominantly occupying the upper region. In contrast, coloring the same embeddings by radiographic finding reveals substantially less distinct clustering. This suggests that progression-aware fine-tuning shifts the representation space toward encoding temporal change rather than primarily finding identity.

\section{Grounding of Progression Semantics: Full Results}
\label{sec:supp_grounding}

\subsection{Setup}

\paragraph{Annotations.}
We use the CheXTemporal gold boxes. Excluding annotations without a current-study box
leaves 1,489 pair--finding--label annotations from 673 image pairs (Improved 354, Stable
560, Worse 381, New 142, Resolved 52; 867 CheXpert, 519 MIMIC-CXR, 103 ReXGradient).
Every model is scored on exactly the same annotations.

\paragraph{Text directions.}
Directions are constructed using each model's own text encoder. Let $s_k$ denote the normalized embedding of progression class $k$ and $s_y$ the target-class embedding. The progression contrast is
\begin{equation}
v_{\mathrm{contrast}} =
s_y - \frac{1}{4}\sum_{k\neq y}s_k.
\end{equation}
Let $Q$ be an orthonormal basis obtained by QR decomposition of three finding-present versus finding-absent text contrasts. The progression-specific direction (\emph{text orth}) is
\begin{equation}
v_{\mathrm{orth}} =
(I-QQ^\top)v_{\mathrm{contrast}}.
\end{equation}
This removes the components aligned with the specified finding-presence contrasts, but does not guarantee removal of every possible presence cue.

We additionally evaluate the full class-sentence direction (\emph{text}), the unprojected progression contrast (\emph{text contrast}), and the mean of the three finding-presence contrasts (\emph{text presence}). In the four-paraphrase variant, each class embedding is averaged over four paraphrases before forming the contrast.

\paragraph{Maps and metrics.}
Each patch is scored by the dot product between its projected patch embedding and the text
direction. Maps are mapped into the 448$\times$448 BioViL-T center-crop frame
(resize 512, crop 448) from each model's own crop geometry and min--max normalized, so all
models are compared against identical box masks. Against the union of current-study boxes
we report pixel AUROC, pointing game (PG), area-matched mIoU, contrast-to-noise ratio (CNR),
and the fraction of map energy inside boxes. Significance uses 2,000 paired bootstrap
resamples over image pairs.

\paragraph{Models.}
Table~\ref{tab:supp_ground_models} lists how each map is produced.
CLIP-style ViTs do not train patch tokens to align with text; we follow the MaskCLIP
approach of passing BiomedCLIP patch tokens through the visual projection head, but its
presence maps are below chance (AUROC 0.378), so BiomedCLIP is reported for completeness
rather than as a meaningful comparison. The \emph{text} and \emph{text contrast} map
variants were not run for external baselines, and the image-encoder preservation analysis
(Appendix Fig.~\ref{fig:supp_grounding_variants}c) applies only to encoders initialized from BioViL-T.
\paragraph{Predictor attribution.}
The main grounding evaluation probes image patches and text embeddings
rather than attribution through the Llama predictor. Its all-to-all
attention does not preserve spatial token correspondence, and direct
predictor-attribution maps yield pixel AUROC 0.517.

\begin{table}[H]
\centering\small
\caption{Grounding models. All maps are mapped into the same 448$\times$448 frame.}
\label{tab:supp_ground_models}
\begin{adjustbox}{max width=\linewidth}
\begin{tabular}{llll}
\toprule
Model & Image input & Patch map & Text encoder \\
\midrule
BioViL-T & prior + current & projected patches, 14$\times$14 at 448 & BioViL-T (CXR-BERT) \\
Prog-JEPA frozen & prior + current & BioViL-T image encoder (unchanged) & Prog-JEPA adapted (LoRA) \\
Prog-JEPA fine-tuned & prior + current & fine-tuned Prog-JEPA image encoder & Prog-JEPA adapted (LoRA) \\
TILA & prior + current & BioViL-T architecture, TILA weights, 14$\times$14 at 448 & TILA \\
BioViL & current & ResNet-50 projected patches, 15$\times$15 at 480 & CXR-BERT-specialized \\
BiomedCLIP & current & ViT-B/16 patch tokens via visual projection, 14$\times$14 at 224 & PubMedBERT (BiomedCLIP) \\
ALTA & [current, 0, prior, 0] & current-slot patches via local projection head & mean of local word embeddings \\
\bottomrule
\end{tabular}
\end{adjustbox}
\end{table}

\subsection{BioViL-T and Prog-JEPA variants}

\begin{table}[H]
\centering\footnotesize
\setlength{\tabcolsep}{3pt}
\caption{Progression-specific grounding (text orth, one class wording) for BioViL-T and
both Prog-JEPA variants. Parenthesized values are differences from BioViL-T with 95\%
bootstrap CIs.}
\label{tab:supp_ground_vljepa}
\begin{adjustbox}{max width=\linewidth}
\begin{tabular}{lccc}
\toprule
Metric & BioViL-T & Prog-JEPA frozen & Prog-JEPA fine-tuned \\
\midrule
Pixel AUROC & .504 & .530 (+.026 [.015, .036]) & .540 (+.036 [.024, .046]) \\
PG & .202 & .235 (+.034 [.009, .059]) & .251 (+.049 [.024, .075]) \\
mIoU & .124 & .150 (+.026 [.020, .032]) & .158 (+.035 [.028, .042]) \\
CNR & .456 & .463 (+.007 [$-$.013, .028]) & .507 (+.051 [.028, .075]) \\
Energy in box & .195 & .199 & .203 \\
\bottomrule
\end{tabular}
\end{adjustbox}
\end{table}

\subsection{Additional Grounding Comparisons}
\label{sec:supp_grounding_variants}

\begin{figure}[H]
  \centering
  \includegraphics[width=\linewidth]{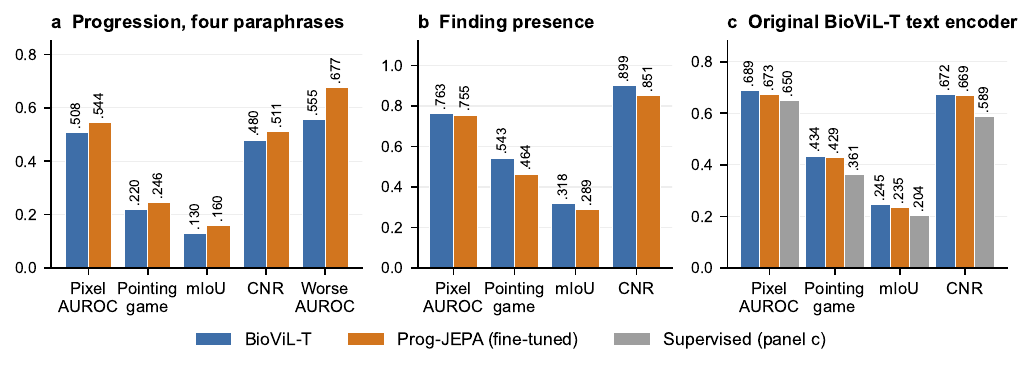}
  \caption{Additional grounding comparisons on CheXTemporal gold. (a) Progression grounding using four paraphrases per class. (b) Finding-presence grounding. (c) Image-encoder preservation using the original BioViL-T text encoder, including the supervised baseline.}
  \label{fig:supp_grounding_variants}
\end{figure}

Figure~\ref{fig:supp_grounding_variants} presents the additional grounding comparisons omitted from the main-paper figure: four-paraphrase progression directions, finding-presence localization, and image-encoder preservation with the original BioViL-T text encoder. All evaluations use the same CheXTemporal gold grounding subset and localization protocol described above.

\paragraph{Four-paraphrase progression grounding.}
Averaging four paraphrases per progression class yields improvements consistent with the single-wording evaluation. Fine-tuned Prog-JEPA increases pixel AUROC from 0.508 to 0.544, pointing-game accuracy from 0.220 to 0.246, mIoU from 0.130 to 0.160, and CNR from 0.480 to 0.511. Worse pixel AUROC increases from 0.555 to 0.677. Thus, the progression-specific improvements persist under alternative class wording.

\paragraph{Finding-presence grounding.}
Fine-tuned Prog-JEPA achieves pixel AUROC of 0.755 compared with 0.763 for BioViL-T, while pointing-game accuracy decreases from 0.543 to 0.464, mIoU from 0.318 to 0.289, and CNR from 0.899 to 0.851. These results show that improved progression-specific grounding does not imply uniformly improved finding-presence localization.

\paragraph{Image-encoder preservation.}
Using the original BioViL-T text encoder, the fine-tuned Prog-JEPA image encoder obtains pixel AUROC of 0.673, pointing-game accuracy of 0.429, mIoU of 0.235, and CNR of 0.669. The corresponding BioViL-T values are 0.689, 0.434, 0.245, and 0.672, respectively. Both outperform the supervised baseline on these metrics (0.650, 0.361, 0.204, and 0.589), indicating that fine-tuning largely preserves the original image encoder's finding-grounding capabilities, despite modest declines relative to BioViL-T.

\subsection{All models}

\begin{figure}[H]
\centering
\includegraphics[width=\linewidth]{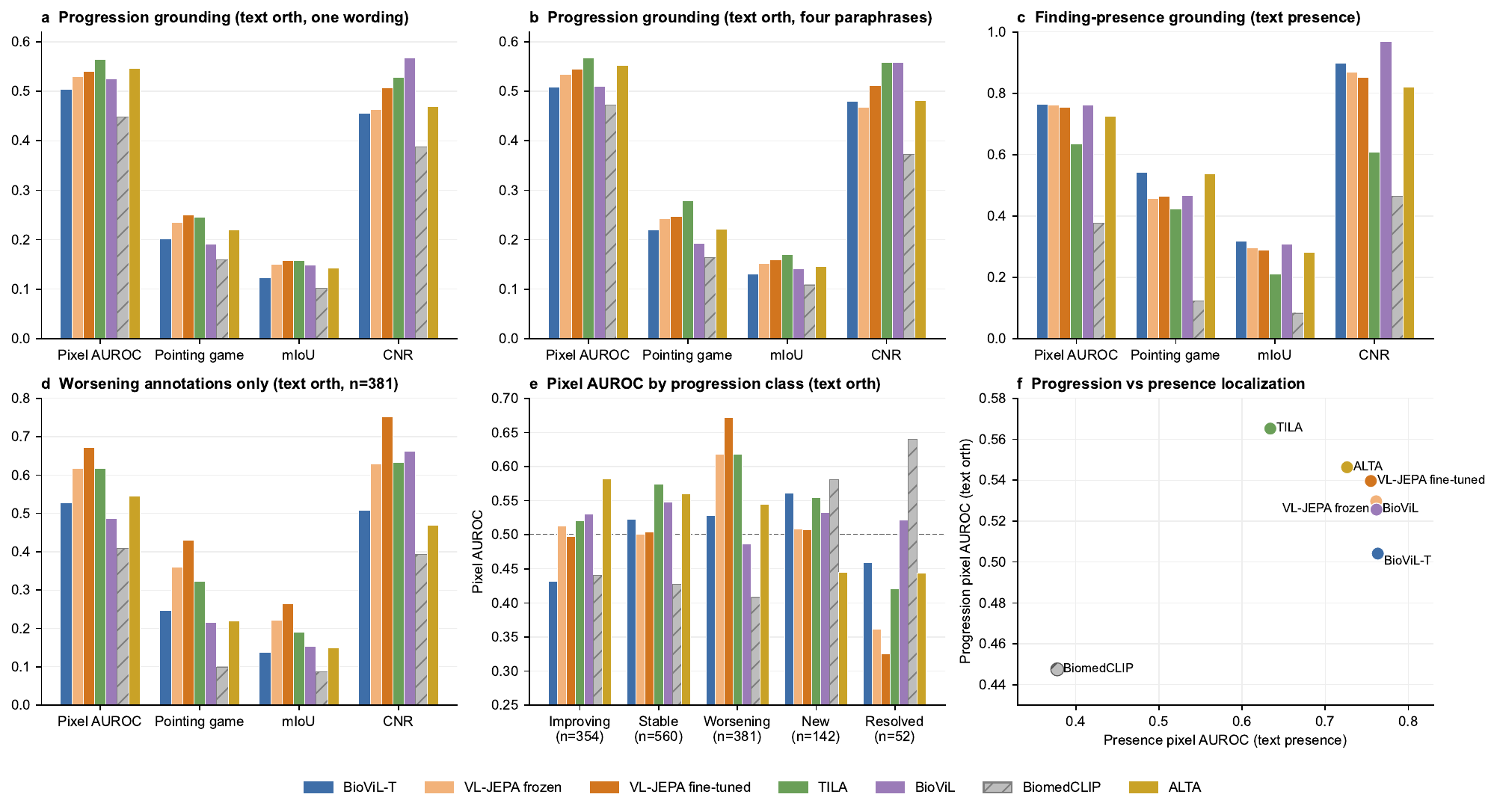}
\caption{Grounding for all seven models. (a,b) Progression-specific grounding (text orth)
with one wording and four paraphrases. (c) Finding-presence grounding. (d) Worse
annotations only. (e) Pixel AUROC by progression class; dashed line is chance.
(f) Progression versus presence pixel AUROC. BiomedCLIP (hatched) produces below-chance
presence maps and is shown for completeness.}
\label{fig:supp_grounding}
\end{figure}

\begin{table}[H]
\centering\footnotesize
\setlength{\tabcolsep}{4pt}
\caption{Progression-specific grounding, one class wording (text orth).}
\label{tab:supp_ground_orth}
\begin{adjustbox}{max width=\linewidth}
\begin{tabular}{lccccc}
\toprule
Model & AUROC & PG & mIoU & CNR & Energy \\
\midrule
BioViL-T & .504 & .202 & .124 & .456 & .195 \\
Prog-JEPA frozen & .530 & .235 & .150 & .463 & .199 \\
Prog-JEPA fine-tuned & .540 & \textbf{.251} & .158 & .507 & .203 \\
TILA & \textbf{.565} & .245 & \textbf{.159} & .528 & \textbf{.206} \\
BioViL & .526 & .191 & .149 & \textbf{.568} & .198 \\
BiomedCLIP & .448 & .159 & .102 & .388 & .185 \\
ALTA & .546 & .220 & .143 & .470 & .197 \\
\bottomrule
\end{tabular}
\end{adjustbox}
\end{table}

\begin{table}[H]
\centering\footnotesize
\caption{Differences from fine-tuned Prog-JEPA (model minus Prog-JEPA fine-tuned), text orth,
one class wording. 95\% bootstrap CIs; bold intervals exclude zero.}
\label{tab:supp_ground_orth_diff}
\begin{adjustbox}{max width=\linewidth}
\begin{tabular}{lcccc}
\toprule
Model & Pixel AUROC & PG & mIoU & CNR \\
\midrule
BioViL-T & \textbf{$-$.036 [$-$.046, $-$.024]} & \textbf{$-$.049 [$-$.075, $-$.024]} & \textbf{$-$.035 [$-$.042, $-$.028]} & \textbf{$-$.051 [$-$.075, $-$.028]} \\
Prog-JEPA frozen & \textbf{$-$.010 [$-$.016, $-$.005]} & $-$.015 [$-$.036, +.005] & \textbf{$-$.008 [$-$.012, $-$.005]} & \textbf{$-$.044 [$-$.057, $-$.030]} \\
TILA & \textbf{+.026 [+.014, +.038]} & $-$.005 [$-$.032, +.021] & +.000 [$-$.007, +.008] & +.021 [$-$.006, +.049] \\
BioViL & \textbf{$-$.014 [$-$.028, $-$.000]} & \textbf{$-$.060 [$-$.086, $-$.034]} & \textbf{$-$.010 [$-$.018, $-$.002]} & \textbf{+.061 [+.036, +.089]} \\
BiomedCLIP & \textbf{$-$.092 [$-$.107, $-$.077]} & \textbf{$-$.091 [$-$.120, $-$.063]} & \textbf{$-$.056 [$-$.065, $-$.048]} & \textbf{$-$.119 [$-$.145, $-$.093]} \\
ALTA & +.007 [$-$.007, +.021] & \textbf{$-$.030 [$-$.059, $-$.001]} & \textbf{$-$.016 [$-$.025, $-$.006]} & \textbf{$-$.037 [$-$.066, $-$.008]} \\
\bottomrule
\end{tabular}
\end{adjustbox}
\end{table}

\begin{table}[H]
\centering\footnotesize
\setlength{\tabcolsep}{4pt}
\caption{Progression-specific grounding, four paraphrases per class (text orth).}
\label{tab:supp_ground_cw4}
\begin{adjustbox}{max width=\linewidth}
\begin{tabular}{lccccc}
\toprule
Model & AUROC & PG & mIoU & CNR & Energy \\
\midrule
BioViL-T & .508 & .220 & .130 & .480 & .194 \\
Prog-JEPA frozen & .535 & .243 & .152 & .468 & .200 \\
Prog-JEPA fine-tuned & .544 & .246 & .160 & .511 & .204 \\
TILA & \textbf{.568} & \textbf{.279} & \textbf{.171} & .558 & \textbf{.206} \\
BioViL & .510 & .193 & .141 & \textbf{.559} & .193 \\
BiomedCLIP & .473 & .164 & .109 & .373 & .189 \\
ALTA & .552 & .221 & .147 & .481 & .198 \\
\bottomrule
\end{tabular}
\end{adjustbox}
\end{table}

\begin{table}[H]
\centering\footnotesize
\caption{Differences from fine-tuned Prog-JEPA, text orth, four paraphrases per class.
95\% bootstrap CIs; bold intervals exclude zero.}
\label{tab:supp_ground_cw4_diff}
\begin{adjustbox}{max width=\linewidth}
\begin{tabular}{lcccc}
\toprule
Model & Pixel AUROC & PG & mIoU & CNR \\
\midrule
BioViL-T & \textbf{$-$.036 [$-$.047, $-$.025]} & \textbf{$-$.026 [$-$.051, $-$.001]} & \textbf{$-$.030 [$-$.037, $-$.023]} & \textbf{$-$.031 [$-$.056, $-$.007]} \\
Prog-JEPA frozen & \textbf{$-$.010 [$-$.015, $-$.005]} & $-$.003 [$-$.025, +.017] & \textbf{$-$.008 [$-$.012, $-$.004]} & \textbf{$-$.043 [$-$.056, $-$.030]} \\
TILA & \textbf{+.024 [+.012, +.036]} & \textbf{+.032 [+.005, +.059]} & \textbf{+.011 [+.003, +.018]} & \textbf{+.047 [+.020, +.074]} \\
BioViL & \textbf{$-$.035 [$-$.049, $-$.020]} & \textbf{$-$.053 [$-$.080, $-$.027]} & \textbf{$-$.019 [$-$.027, $-$.012]} & \textbf{+.048 [+.021, +.075]} \\
BiomedCLIP & \textbf{$-$.072 [$-$.086, $-$.057]} & \textbf{$-$.083 [$-$.112, $-$.054]} & \textbf{$-$.051 [$-$.059, $-$.044]} & \textbf{$-$.138 [$-$.165, $-$.112]} \\
ALTA & +.008 [$-$.007, +.023] & $-$.026 [$-$.055, +.002] & \textbf{$-$.013 [$-$.022, $-$.004]} & \textbf{$-$.030 [$-$.060, $-$.001]} \\
\bottomrule
\end{tabular}
\end{adjustbox}
\end{table}

\begin{table}[H]
\centering\footnotesize
\setlength{\tabcolsep}{4pt}
\caption{Finding-presence grounding (text presence).}
\label{tab:supp_ground_presence}
\begin{adjustbox}{max width=\linewidth}
\begin{tabular}{lccccc}
\toprule
Model & AUROC & PG & mIoU & CNR & Energy \\
\midrule
BioViL-T & \textbf{.763} & \textbf{.543} & \textbf{.318} & .899 & .244 \\
Prog-JEPA frozen & .761 & .458 & .295 & .869 & .248 \\
Prog-JEPA fine-tuned & .755 & .464 & .289 & .851 & \textbf{.250} \\
TILA & .634 & .423 & .211 & .609 & .217 \\
BioViL & .762 & .468 & .307 & \textbf{.970} & .243 \\
BiomedCLIP & .378 & .124 & .083 & .464 & .172 \\
ALTA & .726 & .537 & .281 & .821 & .232 \\
\bottomrule
\end{tabular}
\end{adjustbox}
\end{table}

\begin{table}[H]
\centering\footnotesize
\caption{Differences from fine-tuned Prog-JEPA, text presence. 95\% bootstrap CIs; bold
intervals exclude zero.}
\label{tab:supp_ground_presence_diff}
\begin{adjustbox}{max width=\linewidth}
\begin{tabular}{lcccc}
\toprule
Model & Pixel AUROC & PG & mIoU & CNR \\
\midrule
BioViL-T & \textbf{+.008 [+.002, +.015]} & \textbf{+.079 [+.049, +.109]} & \textbf{+.028 [+.022, +.035]} & \textbf{+.048 [+.027, +.069]} \\
Prog-JEPA frozen & \textbf{+.006 [+.002, +.011]} & $-$.006 [$-$.030, +.018] & \textbf{+.006 [+.001, +.011]} & \textbf{+.018 [+.003, +.031]} \\
TILA & \textbf{$-$.121 [$-$.134, $-$.108]} & \textbf{$-$.041 [$-$.077, $-$.007]} & \textbf{$-$.079 [$-$.089, $-$.070]} & \textbf{$-$.242 [$-$.276, $-$.212]} \\
BioViL & +.007 [$-$.003, +.016] & +.004 [$-$.033, +.037] & \textbf{+.018 [+.009, +.027]} & \textbf{+.119 [+.088, +.147]} \\
BiomedCLIP & \textbf{$-$.377 [$-$.394, $-$.361]} & \textbf{$-$.341 [$-$.377, $-$.307]} & \textbf{$-$.206 [$-$.218, $-$.195]} & \textbf{$-$.388 [$-$.420, $-$.355]} \\
ALTA & \textbf{$-$.029 [$-$.041, $-$.017]} & \textbf{+.073 [+.036, +.109]} & $-$.009 [$-$.019, +.002] & $-$.030 [$-$.067, +.007] \\
\bottomrule
\end{tabular}
\end{adjustbox}
\end{table}

\begin{table}[H]
\centering\footnotesize
\caption{Progression-specific grounding on the 381 Worse annotations (278 pairs),
text orth with one class wording and with four paraphrases (4p).}
\label{tab:supp_ground_worse}
\begin{adjustbox}{max width=\linewidth}
\begin{tabular}{lcccc|cccc}
\toprule
Model & AUROC & PG & mIoU & CNR & AUROC (4p) & PG (4p) & mIoU (4p) & CNR (4p) \\
\midrule
BioViL-T & .528 & .247 & .138 & .508 & .555 & .281 & .154 & .538 \\
Prog-JEPA frozen & .618 & .360 & .221 & .629 & .622 & .360 & .222 & .634 \\
Prog-JEPA fine-tuned & \textbf{.672} & \textbf{.430} & \textbf{.264} & \textbf{.752} & \textbf{.677} & \textbf{.412} & \textbf{.265} & \textbf{.760} \\
TILA & .618 & .323 & .191 & .633 & .640 & .373 & .217 & .687 \\
BioViL & .487 & .215 & .154 & .662 & .486 & .213 & .151 & .670 \\
BiomedCLIP & .408 & .100 & .087 & .394 & .443 & .108 & .097 & .372 \\
ALTA & .545 & .221 & .150 & .470 & .540 & .228 & .147 & .464 \\
\bottomrule
\end{tabular}
\end{adjustbox}
\end{table}

\begin{table}[H]
\centering\footnotesize
\caption{Pixel AUROC by progression class (text orth, one class wording). Parenthesized
values are differences from fine-tuned Prog-JEPA; $^\dagger$ marks a 95\% bootstrap CI that
excludes zero. Annotation counts in the header.}
\label{tab:supp_ground_class}
\begin{adjustbox}{max width=\linewidth}
\begin{tabular}{lccccc}
\toprule
Model & Improved (354) & Stable (560) & Worse (381) & New (142) & Resolved (52) \\
\midrule
BioViL-T & .431 ($-$.067$^\dagger$) & .523 (+.019$^\dagger$) & .528 ($-$.144$^\dagger$) & .561 (+.054$^\dagger$) & .459 (+.134$^\dagger$) \\
Prog-JEPA frozen & .513 (+.015$^\dagger$) & .501 ($-$.003) & .618 ($-$.054$^\dagger$) & .509 (+.001) & .361 (+.036$^\dagger$) \\
Prog-JEPA fine-tuned & .498 & .504 & \textbf{.672} & .507 & .325 \\
TILA & .520 (+.022) & \textbf{.574} (+.070$^\dagger$) & .618 ($-$.054$^\dagger$) & .555 (+.048$^\dagger$) & .421 (+.096$^\dagger$) \\
BioViL & .531 (+.033$^\dagger$) & .548 (+.044$^\dagger$) & .487 ($-$.186$^\dagger$) & .533 (+.026) & .521 (+.196$^\dagger$) \\
BiomedCLIP & .441 ($-$.058$^\dagger$) & .427 ($-$.077$^\dagger$) & .408 ($-$.264$^\dagger$) & \textbf{.581} (+.074$^\dagger$) & \textbf{.640} (+.315$^\dagger$) \\
ALTA & \textbf{.582} (+.084$^\dagger$) & .560 (+.056$^\dagger$) & .545 ($-$.127$^\dagger$) & .445 ($-$.062$^\dagger$) & .444 (+.119$^\dagger$) \\
\bottomrule
\end{tabular}
\end{adjustbox}
\end{table}

\subsection{Findings}

\paragraph{Prog-JEPA versus its initialization.}
Fine-tuned Prog-JEPA localizes progression better than BioViL-T on every metric under both
wordings (Tables~\ref{tab:supp_ground_orth_diff} and \ref{tab:supp_ground_cw4_diff}).
The adapted text encoder alone, with the BioViL-T image encoder frozen, accounts for most
of the pixel-AUROC gain (+0.026 to +0.027); image-encoder fine-tuning adds about 0.01 AUROC
and a large CNR gain (+0.044).

\paragraph{External baselines.}
No model leads on every overall metric. TILA has the highest overall progression pixel AUROC
(+0.026 over fine-tuned Prog-JEPA with one wording) and, with four paraphrases, leads
fine-tuned Prog-JEPA on all four metrics by 0.01--0.05. BioViL has the highest CNR.
ALTA is tied with fine-tuned Prog-JEPA on pixel AUROC and below it on the remaining
metrics with a single wording. Fine-tuned Prog-JEPA has the highest pointing game
with one class wording.

\paragraph{Worse.}
Fine-tuned Prog-JEPA reaches pixel AUROC 0.672 and pointing game 0.430 on Worse
annotations, significantly ahead of every other model on all four metrics with a single
wording (smallest margins: TILA and frozen Prog-JEPA, both $-$0.054 AUROC). With four
paraphrases the same holds except for TILA pointing game ($-$0.039 [$-$0.093, +0.011])
and frozen Prog-JEPA pointing game ($-$0.053 [$-$0.102, +0.000]), which are tied or
borderline (Table~\ref{tab:supp_ground_worse}).

\paragraph{Other progression classes.}
Gains are specific to Worse. Both Prog-JEPA models are near chance on Improved, Stable,
and New (0.50--0.51), where several baselines are significantly higher (ALTA on Improved;
TILA, BioViL, ALTA, and BioViL-T on Stable; BioViL-T, TILA, and BiomedCLIP on New).
On Resolved, where the finding is no longer visible in the current image, both Prog-JEPA
models are well below chance (0.33--0.36) and below every baseline
(Table~\ref{tab:supp_ground_class}).

\paragraph{Presence grounding trade-off.}
TILA's progression advantage comes with a 12-point loss in finding-presence pixel AUROC
(0.634 vs.\ 0.755 for fine-tuned Prog-JEPA) and 0.24 lower CNR. Prog-JEPA keeps presence
pixel AUROC within 0.01 of BioViL-T, although pointing game and mIoU decline
($-$0.079 and $-$0.028; Table~\ref{tab:supp_ground_presence_diff};
Fig.~\ref{fig:supp_grounding}f).

\paragraph{BiomedCLIP.}
BiomedCLIP presence maps are below chance (0.378), indicating that its patch tokens carry
little text-aligned spatial information; its above-chance New and Resolved values should not
be read as genuine localization.

\subsection{Qualitative Grounding Examples}
\label{sec:supp_ground_qualitative}

\begin{figure}[H]
\centering
\setlength{\tabcolsep}{2pt}
\renewcommand{\arraystretch}{0.4}
\begin{adjustbox}{max width=\linewidth}
\begin{tabular}{cccc}
\includegraphics[width=0.23\linewidth,height=0.18\linewidth]{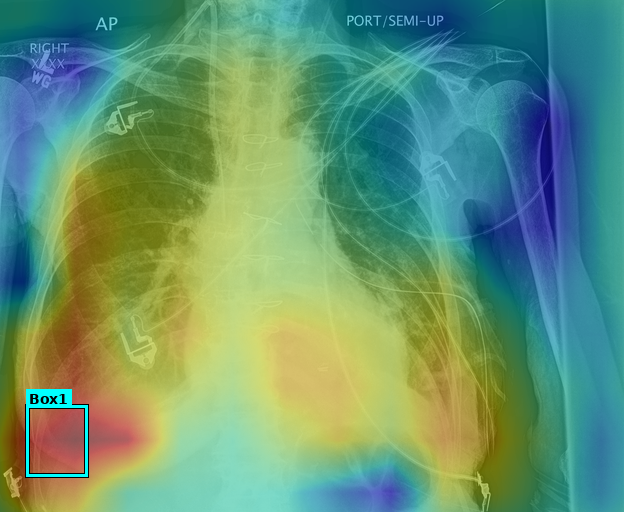} &
\includegraphics[width=0.23\linewidth,height=0.18\linewidth]{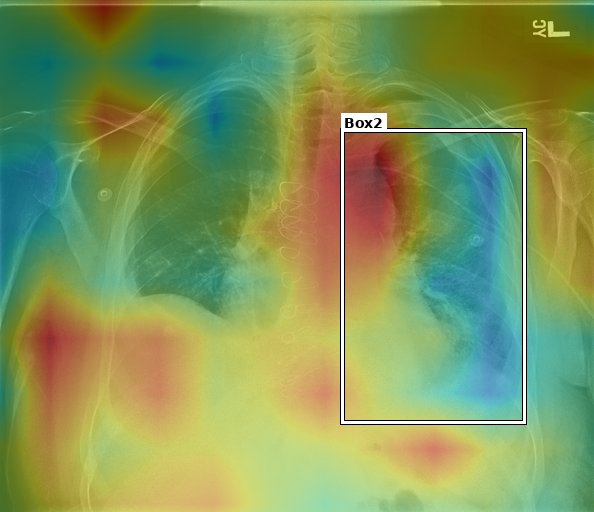} &
\includegraphics[width=0.23\linewidth,height=0.18\linewidth]{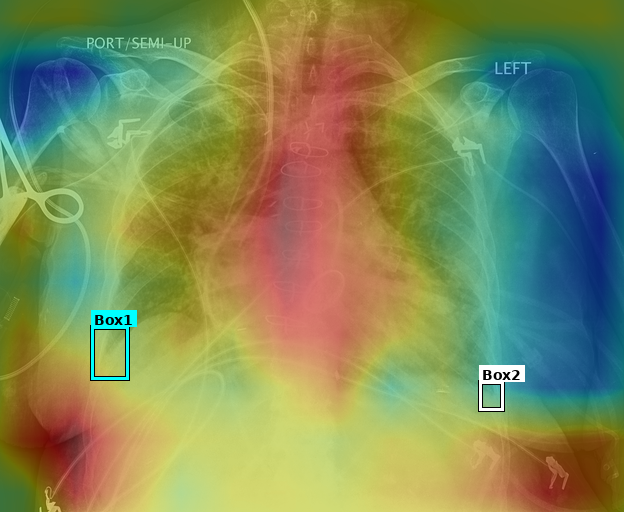} &
\includegraphics[width=0.23\linewidth,height=0.18\linewidth]{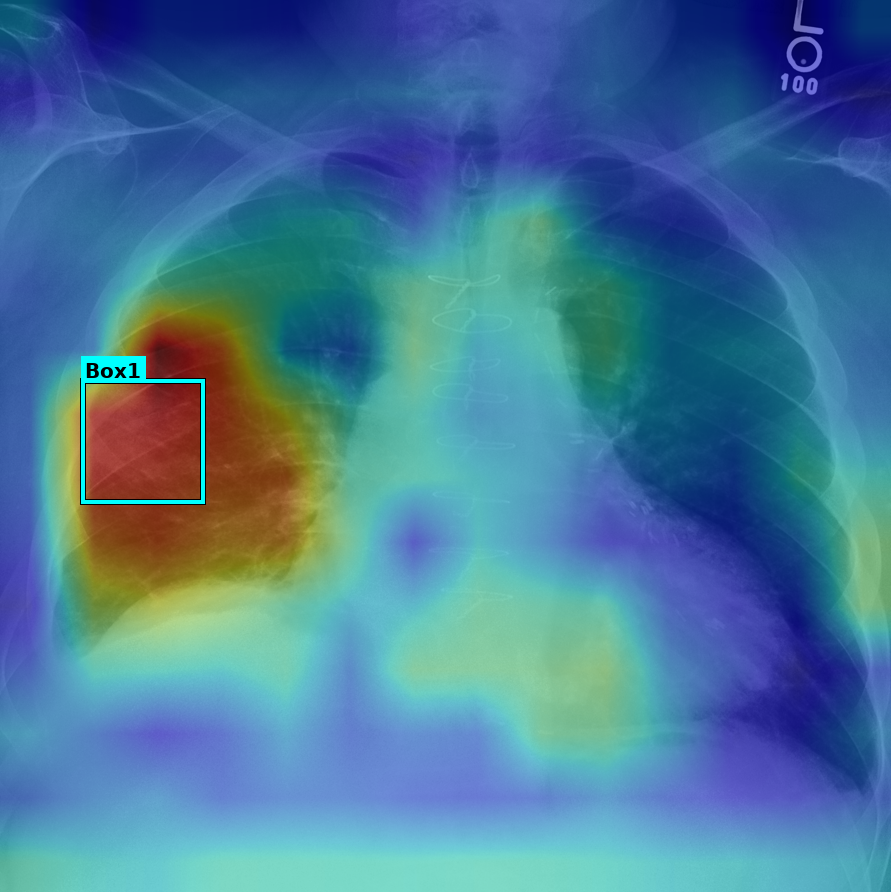}
\end{tabular}
\end{adjustbox}
\caption{
Representative grounding heatmaps across models on the gold dataset. Examples include both high-- and low--performing cases ranked by CNR. Even when global activation strength is high, responses are often spatially diffuse and only coarsely aligned with annotated pathology regions. Peak activations frequently extend beyond the annotated finding or focus on anatomically salient structures rather than pathology--specific regions. Multiple boxes on an image denote different lesions of the same disease.
}
\label{fig:heatmap_examples}
\end{figure}
Representative grounding examples are shown in Figure~\ref{fig:heatmap_examples}. Even in high--CNR cases, activations are often diffuse and only weakly aligned with the annotated pathology region. In lower--performing examples, peak responses frequently occur outside the ground--truth annotation entirely. Qualitatively, models often attend to anatomically prominent regions rather than pathology--specific abnormalities, reinforcing the gap between coarse discrimination and precise temporally grounded localization.

\section{Research Ethics and Data Governance}
\label{sec:supp_ethics}
This study involved secondary analysis of previously
collected, de-identified chest radiography datasets:
CheXpert, MIMIC-CXR, and ReXGradient. The authors did
not recruit or interact with patients, collect new
patient imaging data, or have access to information
that would allow patient identities to be readily
ascertained. No additional IRB review was sought
because the study used existing, non-identifiable
data without patient interaction or intervention.
No additional patient consent was obtained.
All datasets were accessed under their respective
data-use agreements and licensing conditions.
Radiologists contributed expert annotations of
the previously collected images.

\end{document}